%% file: neurips_data_2022.tex
\documentclass{article}

\usepackage[nonatbib, final]{neurips_data_2022}

\usepackage[utf8]{inputenc} 
\usepackage[T1]{fontenc}    
\usepackage{hyperref}       
\usepackage{url}            
\usepackage{booktabs}       
\usepackage{amsfonts}       
\usepackage{nicefrac}       
\usepackage{microtype}      
\usepackage{xcolor}         

\usepackage{graphicx}
\usepackage{amsmath}
\usepackage{amsmath,amssymb}
\usepackage{bbm}
\usepackage{hyperref}
\usepackage{cite}
\usepackage{soul,color}

\usepackage{dsfont}
\usepackage{floatrow}
\newfloatcommand{capbtabbox}{table}[][\FBwidth]
\usepackage{longtable}
\usepackage{listings}

\usepackage[]{quoting}

\usepackage{wrapfig}    

\usepackage{amsthm}

\usepackage{array}          
\usepackage{verbatim}

\title{Communicating Natural Programs \\ to Humans and Machines}

\author{
Samuel Acquaviva\thanks{and $\dagger$ denote equal contributions} \\ MIT
\And
Yewen Pu$^*$ \\ Autodesk Research
\And 
Marta Kryven $\dagger$ \\ MIT
\And 
Theodoros Sechopoulos $\dagger$ \\ MIT
\And
Catherine Wong $\dagger$ \\ MIT
\And
Gabrielle E Ecanow  \\ MIT
\And 
Maxwell Nye \\ MIT
\AND 
Michael Henry Tessler \\ MIT
\And
Joshua B. Tenenbaum\\
MIT
}

\begin{document}

\maketitle

\begin{abstract}
The Abstraction and Reasoning Corpus (ARC) is a set of procedural tasks that tests an agent's ability to flexibly solve novel problems. While most ARC tasks are easy for humans, they are challenging for state-of-the-art AI. What makes building intelligent systems that can generalize to novel situations such as ARC difficult?
We posit that the answer might be found by studying the difference of \emph{language}: While humans readily generate and interpret instructions in a general language, computer systems are shackled to a narrow domain-specific language that they can precisely execute.
We present LARC, the \textit{Language-complete ARC}: a collection of natural language descriptions by a group of human participants  who instruct each other on how to solve ARC tasks using language alone, which contains successful instructions for 88\% of the ARC tasks.
We analyze the collected instructions as `natural programs', finding that while they resemble computer programs, they are distinct in two ways: First, they contain a wide range of primitives; Second, they frequently leverage communicative strategies beyond directly executable codes. We demonstrate that these two distinctions prevent current program synthesis techniques from leveraging LARC to its full potential, and give concrete suggestions on how to build the next-generation program synthesizers.
\end{abstract}

\input{sec_introduction}
\input{sec_comm_and_interp_progs}
\input{sec_larc}
\input{sec_analysis}
\input{sec_synthesis}
\input{sec_related}
\input{sec_conclusion}

\bibliographystyle{unsrt}
{\small \bibliography{main}}

\input{sec_checklist}

\section*{Acknowlegements}
The authors would like to thank Eric Lu for inspiring the wonderful communication game that catalyzed our work.

\input{sec_appendix}

\end{document}

%% file: sec_introduction.tex
\section{Introduction}

\begin{figure}[ht]
    \centering
    \includegraphics[width=0.9\textwidth]{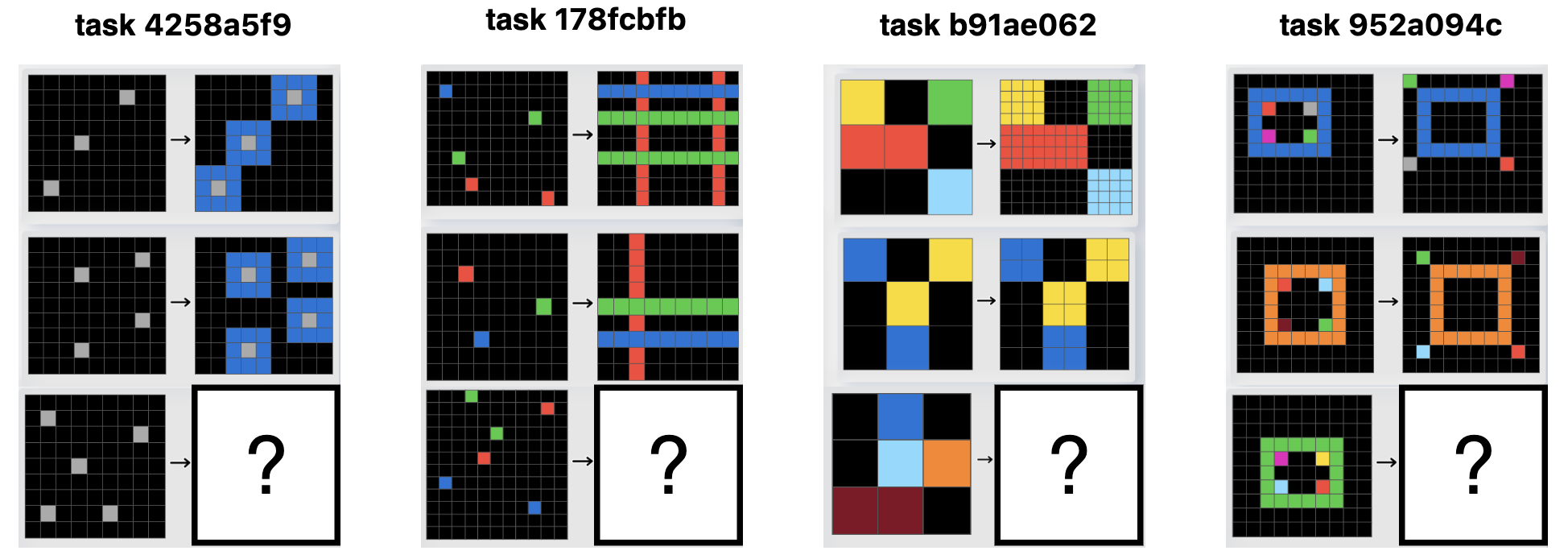}
    \caption{Four ARC tasks, the goal is to correctly infer the unseen output from the given examples.}
    \label{fig:arc_tasks} 
\end{figure}

Humans solve a range of procedural tasks such as cooking, tying shoes, and programming. Although current AI systems achieve super-human proficiency at certain narrowly specified tasks~\cite{silver2017mastering, lerer2020improving}, their reasoning is domain-specific and fails to generalize to novel situations \cite{lake2017building}. The \textit{Abstraction and Reasoning Corpus} (ARC) introduced by \cite{chollet2019measure} presents a set of procedural tasks constructed expressly to benchmark fundamental capacities associated with human general intelligence, including abstraction, generalization, object categories, and procedural analogies ~\cite{chi2014nature,harlow1949formation,lake2017building,lake2020people,bartlett1995remembering,tian2020learning,lombrozo2006structure}. Specifically, ARC requires one to infer a procedure consistent with a small number of abstract input-output examples and apply it to a new input to generate an unseen answer, see Figure \ref{fig:arc_tasks}.

How do we build systems that are capable of solving general, procedural tasks such as ARC? Traditional approaches of program synthesis \cite{parisotto2016neuro,ellis2019write,solar2006combinatorial,devlin2017robustfill} and semantic parsing \cite{artzi2013weakly,ye2020optimal,wang2015building, marzoev2020unnatural, guu2017language, kulal2019spoc} assume the tasks are \textbf{DSL-closed} --  for any task, there exists a program, written in a predefined \emph{Domain Specific Language} (DSL), that solves the task. The ARC benchmark is uniquely \emph{designed} to be \textbf{DSL-open} -- it does not come with a predefined DSL capable of representing its tasks intuitively. This is both reasonable -- most real life tasks, such as cooking and assembling furniture, are DSL-open -- and challenging -- how can one build an intelligent system that can solve tasks from few examples without a DSL? To illustrate, what might a DSL that would allow one to program all the ARC tasks in Figure \ref{fig:arc_tasks} look like? This question is difficult to answer; a recent Kaggle competition found that the best AI systems solve at most 20\% of the tasks, while \cite{johnson2021fast} found that most humans easily solve over 80\% \footnote{Humans were evaluated on a subset of the training tasks; the Kaggle competition used a private test set.}.

Given that humans greatly outperform the best AI systems in solving ARC tasks, studying the human's cognitive processes (for instance, which set of concepts do human use to represent these tasks?) can shed light on how to build similarly intelligent systems. As these thought processes are not observable directly, 
we study \textbf{natural programs} -- instructions that humans give to each other, as a window into these latent cognitive processes. Like computer programs, these instructions can be reliably interpreted (by another human) to produce the intended output. Unlike computer programs, which must be stated in a specific style, natural programs can be stated in any form -- such as verbal instructions or input-output examples -- as long as another human can execute them. In this work, we study a particular form of natural programs, that of \emph{natural language instructions}. We show that analyzing these natural programs -- with explicit comparisons to computer programs -- can both shed light on how humans communicate and interpret procedures \cite{spelke1992origins, itsblockslol, clark1983common, clark1986referring} and inform how one may build AI systems for challenging, DSL-open domains such as ARC.


\begin{figure}[ht]
    \centering
    \includegraphics[width=1.0\textwidth]{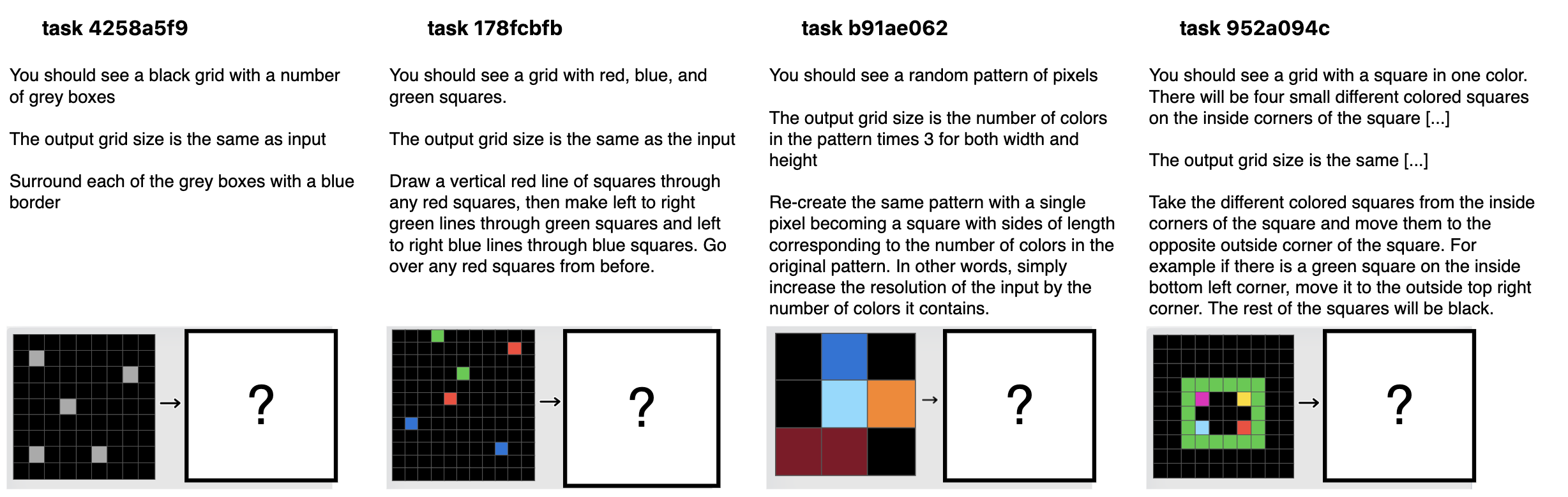}
    \caption{Four LARC tasks, corresponding to those of Figure \ref{fig:arc_tasks}. The goal is to produce the correct output given \emph{only} the language instructions. 88\% of the ARC tasks can be communicated this way. What are some of the communicative strategies used by humans here?}
    \label{fig:larc_tasks} \vspace{-0.25cm}
\end{figure}

We present the \textbf{Language-complete Abstraction and Reasoning Corpus} (LARC) \footnote{\url{https://github.com/samacqua/LARC}} of natural language instructions elicited from a two-player communication game, where 88\% of the ARC tasks can be successfully communicated. LARC tasks are \textbf{language-complete}: The successful instructions contain all the relevant information, even in absence of the original  input-output examples (see Figure \ref{fig:larc_tasks}). This is important in several ways: First, one can use LARC to study how humans use language to communicate abstract procedures, as humans clearly have the capacity to both \emph{generate} and \emph{execute} these natural programs; Second, one can directly see what concepts an intelligent system must be aware of (such as colors and numbers); Third, as people readily generate natural programs, studying them will provide insights on building interactive systems.

We perform \textbf{linguistic analysis} on LARC, finding that humans readily leverage algorithmic concepts without being explicitly instructed to do so. These concepts range from domain general ones, such as loops, to domain-specific concepts such as flood-fill. However, natural programs in LARC are distinct from typical computer programs in two ways: (1) natural programs use a much wider range of concepts compared to a typical DSL; (2) natural programs contain clarifications and validations in greater quantity than directly executable procedures. We apply standard \textbf{program synthesis} algorithms on LARC, finding that while existing approaches can benefit from the additional language annotations, the two aforementioned distinctions pose significant challenges to standard program synthesis approaches.
We conclude by providing concrete suggestions on how to build the next generation program synthesizers.


%% file: sec_comm_and_interp_progs.tex
\section{Communicating and Interpreting Programs}

\begin{figure*}[!t] 
    \centering
    \includegraphics[width=0.98\textwidth]{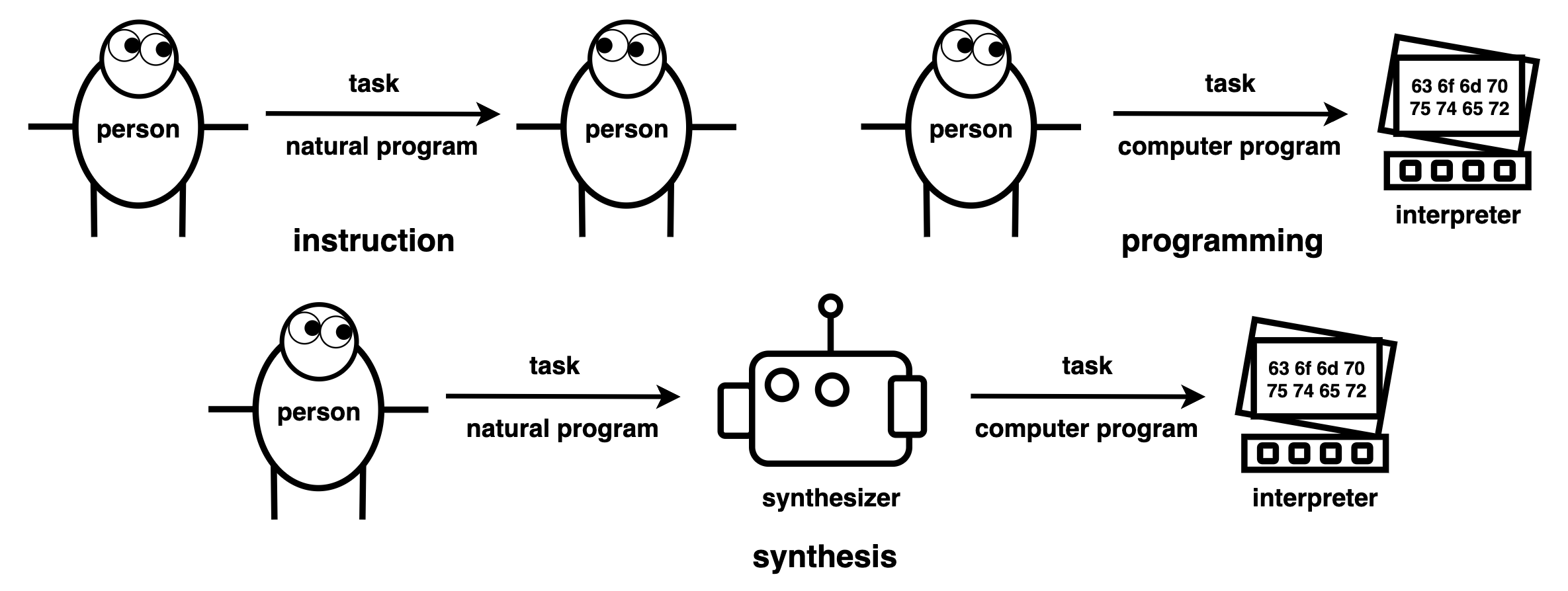}
    \caption{Three kinds of ``programs'': instruction (top-left), programming (top-right), synthesis (bot).}
    \label{fig:three_interactions}
\end{figure*}

In programming, a \emph{programmer} constructs a \emph{program} in a suitable language, which is then executed on an \emph{interpreter}, producing a behaviour. For instance, a person can \emph{instruct} another person to carry out a certain task (Fig. \ref{fig:three_interactions} top-left),
or directly \emph{program} a machine to solve tasks using code (Fig. \ref{fig:three_interactions} top-right). A program synthesizer takes in an instruction, and reformulates it as code, insulating the person from the programming process  (Fig. \ref{fig:three_interactions} bot). We treat all three as acts of \emph{programming}. 

How do we build systems that can be communicated naturally to solve challenging tasks?
Typically, one follows a ``DSL-first'' approach, where one first defines a programming language and builds a corresponding interpreter capable of executing programs written in this language. Then, one naturalizes the initial DSL using synthesis, allowing end-users to describe tasks using natural language \cite{artzi2013weakly,artzi2014learning,ye2020optimal,wang2015building, wang2016learning, marzoev2020unnatural}, or by giving examples  \cite{ellis2019write,solar2006combinatorial,pu2020program}. While this DSL-first workflow has yielded impressive results, the DSL itself is also a single point of failure. It is difficult to design DSL with the right \textit{scope}, so that it both expressive and non-redundant \cite{gulwani2015inductive,bruce2020jshrink,soto2021comprehensive}. One must ensure that the DSL \emph{aligns} reasonably to human instructions \cite{liang2016learning,shin2021constrained}, while simultaneously being \emph{efficient} when used by the synthesizer \cite{ellis2019write, ellis2020dreamcoder}. These challenges may explain why ARC, and other DSL-open domains (where procedural tasks are given \emph {in the absence} of a narrow DSL), are difficult to tackle.

In this work, we adopt the Wizard-of-Oz approach \cite{kelley1984iterative, budzianowski2018multiwoz, wen2016network} by using a human as an interpreter of natural language instructions (Fig \ref{fig:three_interactions} top-left). We define a \textbf{natural program} as instructions constructed by a person that can be interpreted by another person to produce a specific output. This program is \textit{natural}--it can be understood by speakers of the language\footnote{language here is to be understood loosely as any medium of communication between people} without a prior consensus--but behaves as a \textit{program}, in that it produces a definitive output, which can be unambiguously checked for correctness. For instance, the original ARC \cite{chollet2019measure} tasks are natural programs: Given a program consisting of input-output examples, a fellow human can readily interpret this program to produce an output on a new input, which can be checked for correctness.
By starting with (linguistic) natural programs, one can directly observe the set of concepts and strategies necessary to master a domain (such as ARC), without committing to a specific interpreter.

\begin{figure}[ht]
    \centering
    \includegraphics[width=0.7\textwidth]{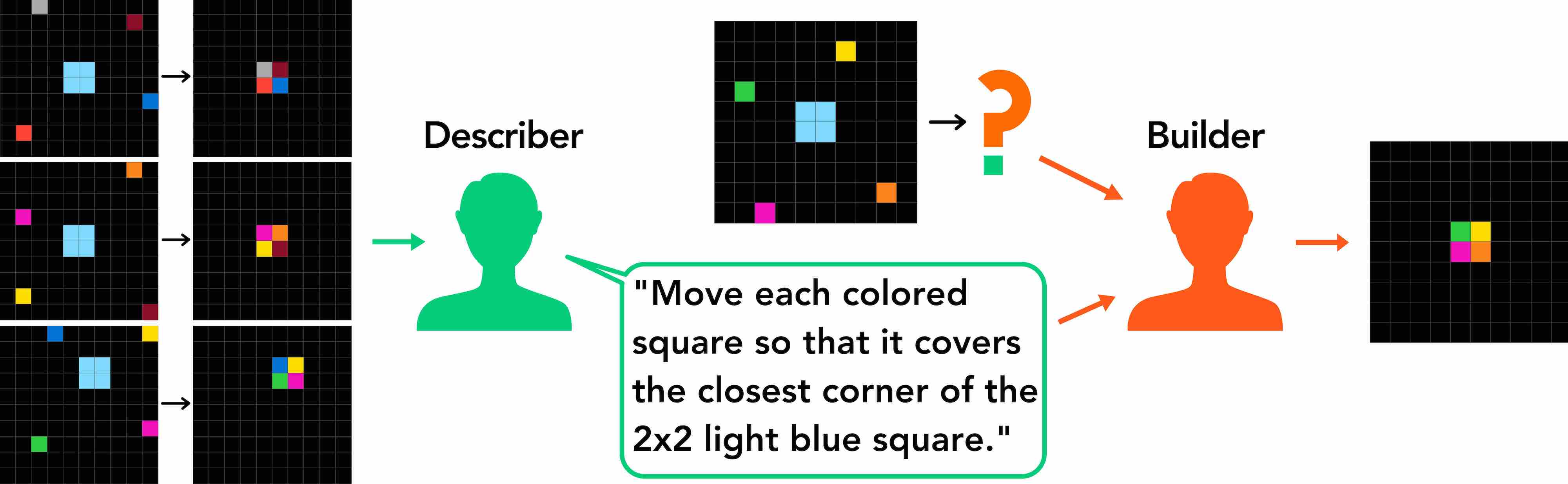}
    \caption{a \textbf{describer} instructs a \textbf{builder} how to solve an ARC task using a natural program}
    \label{fig:game} \vspace{-0.25cm}
\end{figure}

%% file: sec_larc.tex
\section{LARC: Language-complete Abstraction and Reasoning Corpus}
\label{sec-larc}

\begin{figure*}[t]
\centering
    \includegraphics[width=0.85\textwidth]{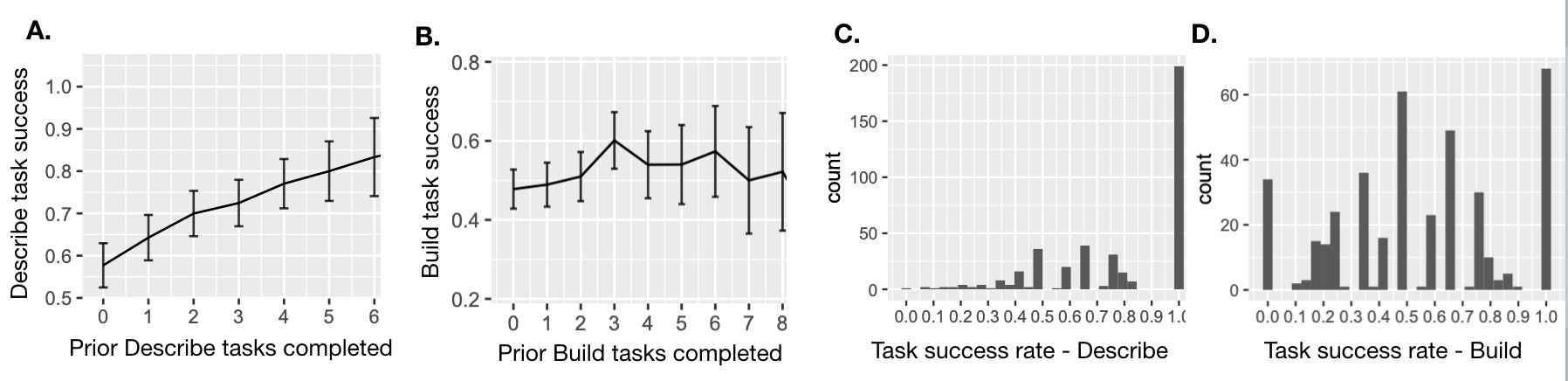}
    \caption{ { \small 
    \emph{A.} Describer improves at verifying their own descriptions as a they describe more tasks. 
    \emph{B.} Builders do not improve at constructing the correct outputs as they build more tasks (likely due to having no control over the qualities of their given descriptions).
    \emph{C.} Rate of describers verifying their own descriptions (avg 75\%). \emph{D.}  The rate of builders constructing the correct output, (avg 50\%).}}
    \label{fig:results2} \vspace{-0.2cm}
\end{figure*}

We present a dataset that augments the original ARC tasks from \cite{chollet2019measure} with \emph{language-complete} instructions: they can be demonstrably interpreted by other humans to correctly produce the intended outputs without any additional contexts (i.e. in the absence of the original input-output examples). Thus, LARC tasks (Fig \ref{fig:larc_tasks}), like their counterparts in ARC, meet the definition of natural program while containing only natural language descriptions. To collect this dataset, we introduce a \emph{communication game}: human describers produce linguistic instructions from the given input-output examples of ARC, these instructions are then interpreted by human builders (in the absence of the original input-output) on a new instance of the same task (Fig. ~\ref{fig:game}). We deployed this experiment using a novel 
bandit algorithm to efficiently collect verifiable natural programs.
The final dataset augments 88\% of the original ARC tasks (354/400) with at least one verifiable \textit{natural program} description that could be successfully interpreted by another human participant to solve the task. Fig.~\ref{fig:results2}(C-D) shows the distribution of success rates for participants acting as describers and builders over time.


\subsection{Human annotation details}
We recruited 373 subjects via Amazon Mechanical Turk who were paid for 45 minutes of work. Fifty individuals were excluded for failing to complete the task, so the final analysis included 323 subjects. The study was approved by our institution's Institutional Review Board, did not collect personally identifiable information, and did not pose risks to participants. Subjects were paid \$6.00 and a \$0.25 bonus for every successful communication. Subjects averaged 5.5 communications, bringing their expected hourly wage to \$9.83. For interface and consent form see Appendix \ref{sec:consent_form}. \footnote{see \url{https://arxiv.org/abs/2106.07824} for full paper with appendix attached at the end}

\subsection{Two-player communication game}\label{sec:proc}

For each task, a participant may be assigned one of two roles: a \textbf{describer} or a \textbf{builder}. 
The describer plays the role of a \emph{human synthesizer}, who reformulates input-output examples (of ARC) to natural language descriptions. 
The builder plays a role of a \emph{human interpreter}, who must construct the correct output on a new input without access to the original examples (Fig \ref{fig:game}).
The description is structured into three sections to incentivize consistency: (1) what the builder should expect to see in the input, (2) the output grid size, and (3) what the builder should do to create the output (Fig~\ref{fig:larc_tasks}). 
After the description was submitted, we verify the describer's own understanding by asking them to build it, and discarding the submission if the describer fails.
The describer was shown all previous verified descriptions for a task, alleviating challenge of solving the task from scratch.
Builders construct/draw the output using actions defined in ARC, such as \texttt{paint(color,x,y)}, \texttt{copy/paste}, and \texttt{floodfill}. All drawing sequences are recorded and can be played back.

\subsection{The Bandit Algorithm for Data Collection}\label{sec-collection}

Collecting valid linguistic natural programs requires significant human efforts: For each task (of varying difficulties), natural programs must first be \emph{proposed} by a number of describers, and then \emph{validated} by a number of builders, where both can make mistakes. Thus, A naive data-collection process that simply collects a fixed number of descriptions and builds per task will be expensive. To address this challenge, we formulate the following bandit problem: \emph{multi-bandit} -- each of the 400 ARC tasks is a different bandit; \emph{infinite-arm} -- given a task, each natural language description (there are infinitely many) is a different arm; \emph{best-arm identification} -- once a natural program is proposed, we must validate it. 
We develop a novel bandit algorithm (Appendix \ref{sec:bandit_algo}) to solve this problem, as to our best knowledge, no known bandit algorithm can be directly applied. 
For each MTurk participant, our bandit algorithm dynamically allocates a set of describing and building efforts for their session. As a result, the LARC dataset was annotated for \$3667, whereas a naively collecting 20 annotations per task would cost at least \$10,800.

\begin{figure*}[t]
\centering
    \includegraphics[width=0.9\textwidth]{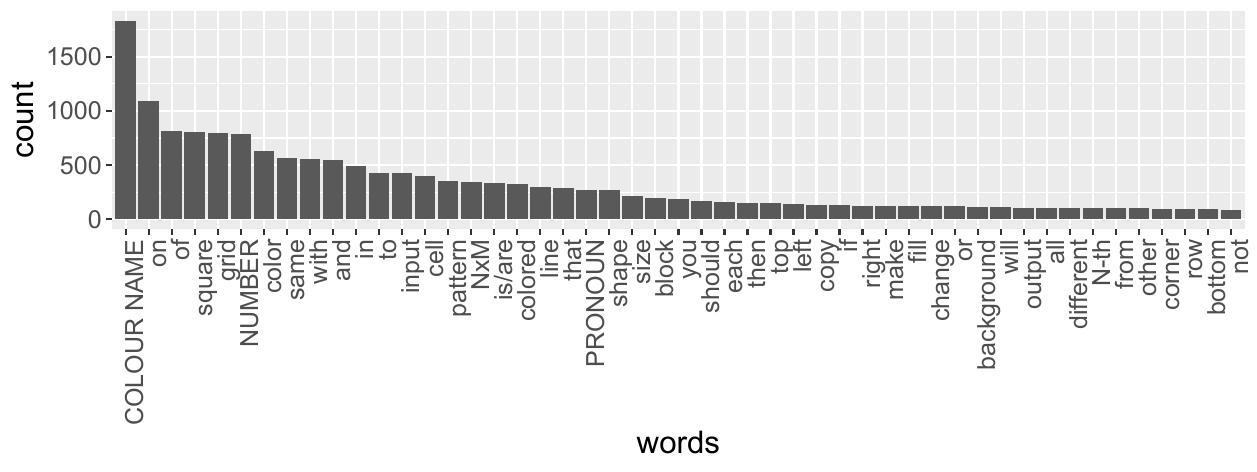}
    \caption{ { \small Words used in successfully built descriptions, sorted by their frequency in the corpus (total 642 unique words). The words were singularized. Colors names, numbers, and pronouns were grouped together.} }
    \label{fig:results3} \vspace{-0.5cm}
\end{figure*}

%% file: sec_analysis.tex
\section{Communication Strategies in Natural Programs}
\label{sec-analysis}

What are some strategies humans use to produce robustly interpretable instructions? 
To answer this question, we curate a \textit{linguistically tagged dataset} of tagged phrases from successful descriptions under the lens of \emph{computer programs}. We annotate these phrases with \textit{tags} corresponding to general concepts from algorithms and \emph{core knowledge} \cite{spelke2007coreknowledge}.
In total, we manually label 532 randomly sampled phrases (22\% of the phrase corpus) using 17 conceptual tags (in which multiple tags can be applied to each phrase); Figure ~\ref{fig:tags}A. shows a frequency of these tags.
For details see Appendix \ref{sec:tagging_scheme}.

\subsection{Similarities of Computer and Natural Programs}

\paragraph{General Algorithmic Concepts} LARC contains algorithmic concepts similar to those found in a typical programming language (i.e. python). For instance, \textbf{tag\_logic} is a boolean check (i.e. ``the box is blue''), \textbf{tag\_array} references a set of similar objects (i.e. ``you should see four red shapes''), and \textbf{tag\_loop} is similar to loops (``keep going until ''). Humans generate these concepts without being directly instructed to do so, suggesting that humans reason about ARC tasks algorithmically.

\begin{figure*}[t]
\centering
    \includegraphics[width=0.9\textwidth]{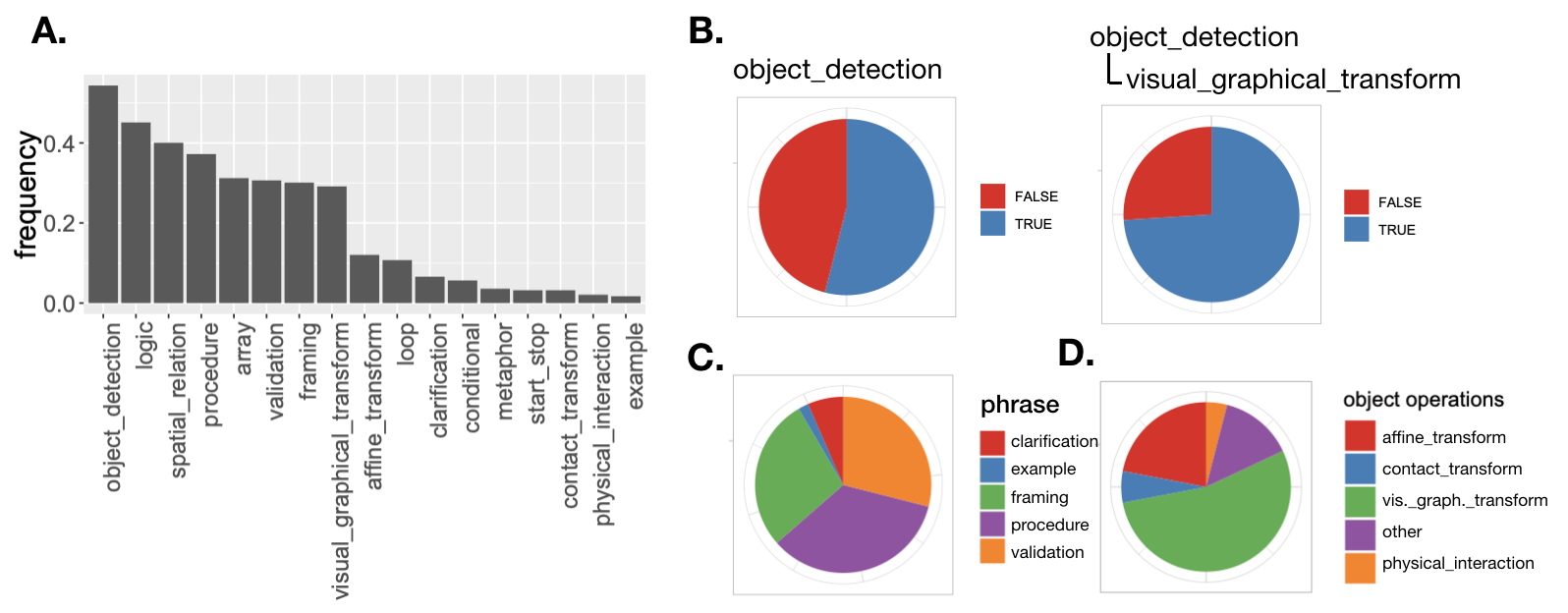}
    \caption{ { \small \textbf{A.} The frequencies of all tags occurring in human phrases. Each phrase can have multiple tags. 
    \textbf{B.} More than half of the phrases described objects, of which, 75\% described spatial relations. 
    \textbf{C.} Relative frequencies of code (procedures) against non-code (example, framing, clarification, validation).
    \textbf{D.} Relative frequencies of core knowledge topics in phrases that referenced objects.} }
    \label{fig:tags} \vspace{-0.5cm}
\end{figure*}

\paragraph{Domain Specific Concepts} Similar to a computer DSL, LARC contains concepts that distinguish it from other domains. We focus on the object system of core knowledge \cite{spelke2007coreknowledge}, defined by \emph{cohesion, persistence}, and \emph{influence via contact}, which the ARC corpus was explicitly designed to leverage. We find about half of the phrases referenced \textbf{objects}, and three quarters of these described spatial relations (Fig.\ref{fig:tags}B). Majority of operations on objects (Fig.\ref{fig:tags}D) are \textbf{visual\_graphical\_transform} whereas only 5\% of are \textbf{physical\_interaction}. Presumably, graphical transformations are easier to represent in the  input-output format of ARC.

\subsection{Differences of Computer and Natural Programs}

We outline two (related) ways natural programs differ from computer programs. First, instead of using a narrow DSL with few primitives, natural programs use a large, diverse set of primitive functions. Second, instead of stating a precise procedure verbatim, natural programs rely on a range of additional strategies to ensure that they can be interpreted precisely.

\paragraph{Natural Programs Invoke a Large Number of Concepts}
Since LARC is language-complete, analyzing the words used in LARC serves as a good proxy for the underlying concepts
present in the ARC domain. Similar to \cite{johnson2021fast}, we find that humans use a wide range of concepts (Fig \ref{fig:results3}). This is a testament of the general capabilities of the \emph{human interpreter}: the describers readily invoke these concepts from the builders, with the confidence that they can be correctly interpreted.
Given a large number of concepts, effectively indicating the set of relevant concepts (for a given task) becomes nontrivial: While human describers and builders can make use of generic word such as `bump into', computer programmers must be extremely careful in selecting the exact concept using a precise language (i.e. \texttt{move\_until\_touches\_block}). 

\paragraph{Natural Programs Communicate Information Beyond Procedures}
We study the relative frequencies of \emph{directly executable} commands, \textbf{tag\_procedure}, in contrast to not directly executable meta information such as \textbf{tag\_framing} -- comments about which concepts are relevant, \textbf{tag\_validation} -- checks to ensure correct execution, and \textbf{tag\_clarifications} -- restating the same procedure in different words. The most striking finding is that procedure, framing, and validation occur at roughly the same frequency (see Fig.\ref{fig:tags} C). In contrast, only 14\% of the codes are commented \cite{huang2020does}. 

The high frequency of framing tags suggests that describers anticipate the large number of concepts that the builder can operate over, and carefully \emph{frame} the instruction to invoke the appropriate ones. 
The describer often assumes the directly executable portion (i.e. tag\_procedure) as inherently \emph{ambiguous}, as suggested by frequent use of tag\_validations and tag\_clarifications following these procedures. Specifically, validation gives a check to the builder to test if their current interpretation is correct. Clarification amends the initial ambiguous explanation with another explanation, narrowing the number of possible interpretations. These are evidences that, unlike communication in computer programs over a \emph{narrow and unambiguous} DSL, communication in natural programs are fundamentally \emph{expressive yet ambiguous}, requiring extra efforts to maintain precision.

%% file: sec_synthesis.tex
\section{Executing Natural Programs using Program Synthesis}\label{sec-synthesis}

We evaluate whether current DSL-first program synthesis methods (Fig \ref{fig:three_interactions}, bot) can execute natural programs as well as humans do. We consider three kinds of natural programs: (1) Input-output examples from the original ARC corpus (IO); (2) IO in conjunction with successful language instructions in LARC (IO+NL); And  (3) language alone (NL-only) -- same as the MTurk builder task. 

\subsection{Program Synthesis} 

In (symbolic) program synthesis \cite{solar2008program,devlin2017robustfill,guu2017language}, the synthesizer takes in a natural program, and reformulates it as code over a DSL to be executed. We have manually crafted a DSL based loosely on the concepts present in the LARC corpus and built its corresponding interpreter (see Appendix \ref{sec:larc_dsl}) \footnote{this is a tremendous engineering effort, consisting of 103 primitives compared to 33 of SCONE}. We present our best synthesis results here. For additional models (using a CNN encoder, a sequence decoder \cite{guu2017language}) see \ref{sec:other_experiments}. Preliminary studies with \emph{codex} and \emph{clip} see \ref{sec:copilot} and \ref{sec:clip}.

\paragraph{Generate and Check Using IO}
If the given natural program contains IO examples, the standard symbolic program synthesis approach \cite{solar2006combinatorial,devlin2017robustfill} follows the \emph{generate and check} strategy. Let $natprog$ be a natural program, the synthesizer returns programs $prog$ from a DSL from the following distribution:
$$
P_{synth}(prog | natprog) \propto P_{gen} (prog | natprog) \mathds{1}[prog \vdash IO]
$$
$P_{gen}$ is the generative distribution: given a natural program, it proposes program $prog$ from the DSL. $\mathds{1}[prog \vdash IO]$ is the checker: it validates $prog$ by executing it on the interpreter, ensuring that $prog(x)=y$ for all input-output pairs $(x,y) \in IO$. The key strength of this approach lies in its generalizability:
If a proposed program can be checked against all IO examples, it is very likely to generalize to an new instance of the same task due to the inductive bias of the DSL.

\paragraph{Generation Models} 
Our $P_{gen} (prog | natprog)$ generates programs in two parts: a neural model outputs a tree bigram over the grammar of the DSL \cite{lari1990estimation}, then a dedicated Ocaml enumerator deterministically enumerates programs from a probabilistic context free grammar fitted to this bigram distribution in decreasing probability \cite{ellis2020dreamcoder}.
For simplicity, we report results of unconditioned generators $P_{gen}(prog)$ (i.e. a fitted prior) when language is absent, and language-conditioned models $P_{gen}(prog|NL)$ when language is present. This way, we can use the same $P_{gen}(prog|NL)$ model for both IO+NL and NL-only tasks in the test set, as it does not depend on IO. Similar to  \cite{dechter2013bootstrap,wong2021language}, we first bootstrap our generative models with 10 ``seed'' programs, discovered uninformed enumeration. 

\paragraph{Leveraging Language} We use a pre-trained model (T5, \cite{raffel2019exploring}) to represent language by taking an average of its encoded tokens.
To encourage the learning of compositional relationships between language and program, we use \textbf{pseudo-annotation}, similar to recent methods that have leveraged synchronous grammars \cite{marzoev2020unnatural,jia2016data,shin2021constrained,wong2021language}. First, we provide linguistic comments for each primitive function in the program DSL (e.g. \texttt{flood\_fill(color)} with \textit{fill with the color}). Then, during training, we obtain additional paired language and program examples by substituting primitives of artificial programs with their corresponding comments \footnote{
for instance, \texttt{(lambda (to\_original\_grid\_overlay (remove\_color(grid\_to\_block x) yellow) false)) }  becomes
\textit{place block on input grid remove color from block yellow} }.
For more examples see Appendix \ref{sec:larc_dsl}.

\paragraph{Distant Supervision}
LARC, similar to SCONE \cite{long2016simpler}, falls under the challenge of distant supervision: each training task only contains the correct output, but not the ground-truth program responsible for generating it. We adopt the iterative approach used in  \cite{dechter2013bootstrap,ellis2020dreamcoder,wong2021language,guu2017language} to \emph{discover} suitable programs during the training phase, by alternatively (1) generating a large sample of programs using $P_{gen}$ and (2) fitting a better $P_{gen}$ from good programs in the generated samples.

\subsection{Results}

\begin{figure}
\begin{floatrow}
\capbtabbox{%
\begin{tabular}{ccc}

 \multicolumn{3}{c}{training tasks discovered} \\ \hline
   & no-pseudo & pseudo \\
\hline
IO & 15 / 200 & - \\
IO + NL & 13 / 200 & \textbf{21 / 200} \\ \hline
& & \\
& & \\
 \multicolumn{3}{c}{testing tasks solved} \\ \hline
   & no-pseudo & pseudo \\
\hline
NL-only & 1 / 183 & 0 / 183 \\
IO & 18 / 183 & - \\
IO + NL & 16 / 183 & \textbf{22 / 183} \\ \hline
\end{tabular}
}{%
 \caption{Executing different kinds of natural programs (IO -- Input-output examples from the original ARC corpus, IO+NL -- IO in conjunction with successful language instructions in LARC, NL-only -- same as the MTurk builder task) using program synthesis. Here, "pseudo" means the NL training has been pre-trained on generated synthetic language to code pairs. Train tasks discovered under distant supervision (top). Test tasks solved (bot).  }%
 \label{tab:result}
}
\ffigbox{%
\centering
\includegraphics[width=0.53\textwidth]{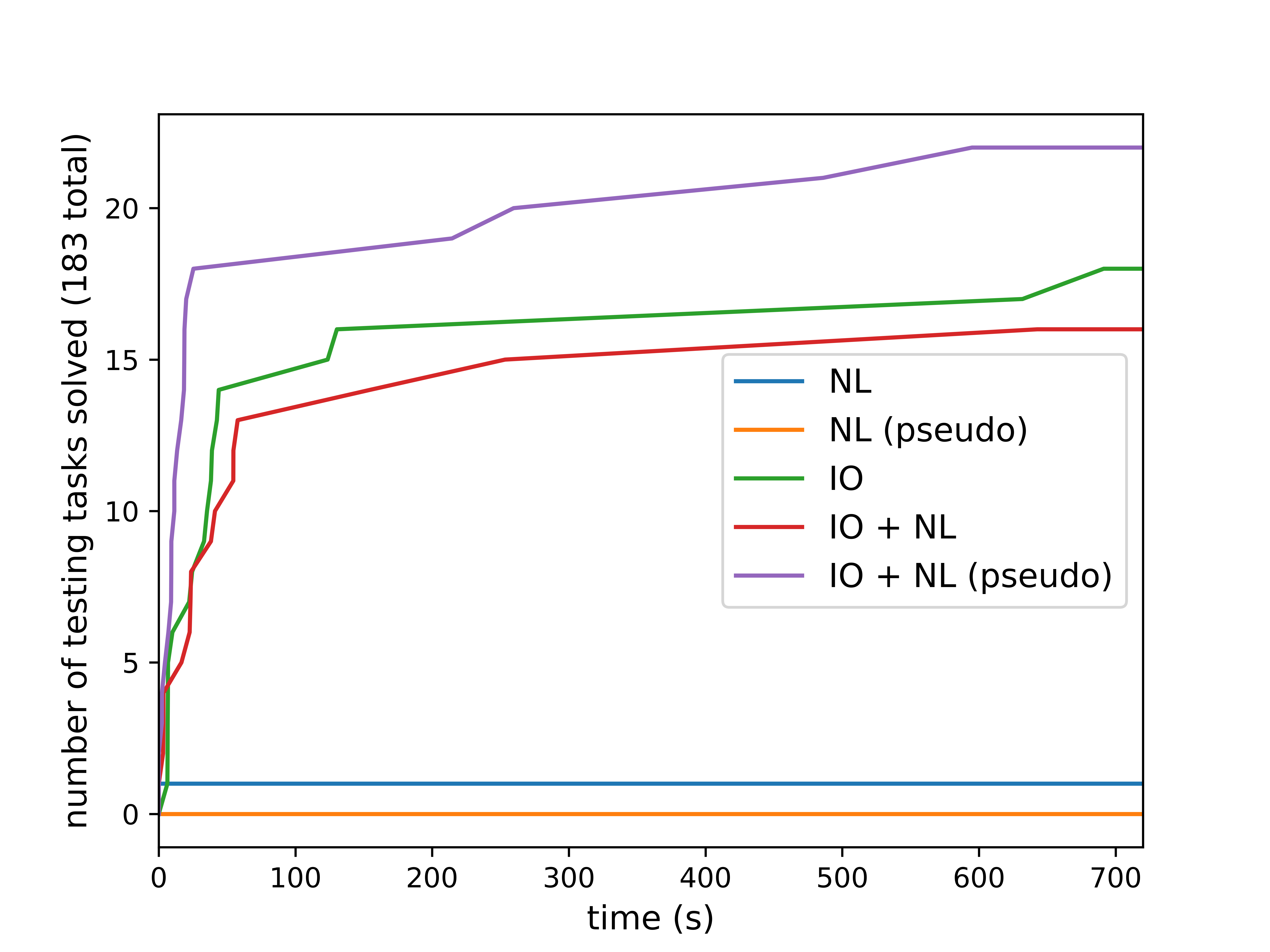}
}{%
  \caption{Number of test tasks solved for the three kinds of natural programs, IO, NL, IO+NL, with and without pseudo annotations, as a function of enumeration time. There are error bars as the bigram enumerator is deterministic. It is possible (but not likely) that re-training these models will have an effect due to the randomness of sampling pseudo-annotated programs. All models vastly under performs when compared to a human, but natural programs consisting of NL+IO fairs best.}%
  \label{fig:synth_results}
}
\end{floatrow}
\end{figure}

We split the 400 tasks into 200 training tasks (with or without valid language descriptions) and 183 testing tasks (the remaining 200 filtered for having valid language deceptions). We then train the models for 10 hours each using iterative learning. We test on the 183 test tasks by first using the neural model to propose a bigram per task, then enumerating the bigram for 720 seconds. We keep the top-3 most likely programs that also satisfy the IO examples if the natural program contains IO. We then check if any of the top 3 programs satisfies test input-output. See Table \ref{tab:result} and Figure \ref{fig:synth_results}. Overall, we conclude that while language definitely helps current approaches, the overall results (best 12\%) are still comically bad.

\paragraph{Quantitative Findings}
IO+NL+psuedo performs best, solving 22/183 of the testing tasks. 
We believe this due to psuedo-annotation being able to generate an infinite number (albeit low quality) of artificial NL-prog pairs. We note that having the ability to \emph{check} if a proposed program is correct under IO is \emph{crucial} for the success of current program synthesizers, with no more than 1 task solved with NL-only. Like the validation phrases in LARC, the input-output examples in IO serve as a form of \emph{validation} for the enumerative synthesizer. This finding corroborates with \cite{austin2021program}.

\begin{figure*}[h] 
    \centering
    \includegraphics[width=\textwidth]{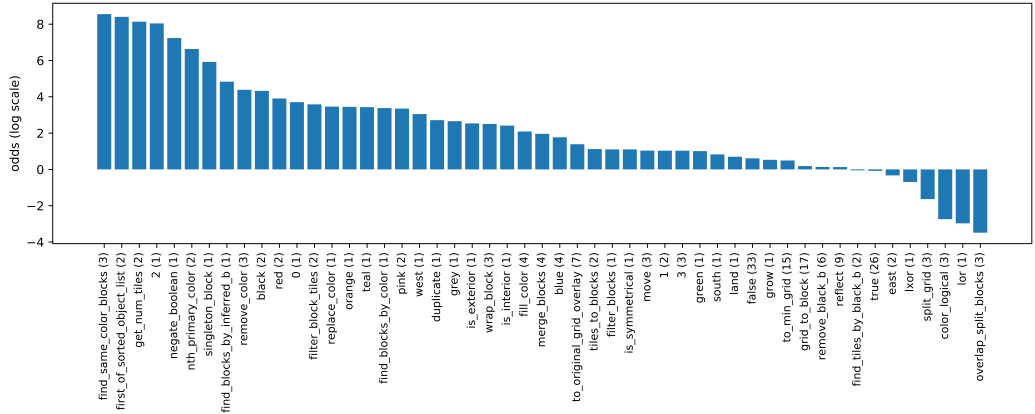}
    \caption{Relative odds of using the correct primitive for a task, language-conditioned vs unconditioned generation. Number in parenthesis denotes the total number of times a primitive is used.}
    \label{fig:primitive_odds}
\end{figure*}

\paragraph{Qualitative Findings}
We investigate in what way does language affect synthesis. For each primitive in our DSL, we ask how many times more likely is it going to appear in correct programs generated with the language-conditioned bigram vs the unconditioned one. We plot this ratio on a log scale for all primitives that were used in ground-truth programs, see Figure \ref{fig:primitive_odds}. We note that for most of the frequently used primitives, the language-conditioned generator is more likely to generate the correct primitives than the unconditioned generator. 

\subsection{Challenges}

The biggest challenge is \textbf{scoping}. Since LARC is DSL-open, we were in a vicious cycle of constantly adding more primitives and refactoring the DSL. Even now, we cannot guarantee our DSL can represent all LARC tasks. Second challenge is \textbf{referencing}: with 103 primitives, selecting the relevant primitives becomes crucial \footnote{if we can magically select $10$, the search space is $10^5$ instead of $103^5$ for a program of length $5$}. Finally, current NL-to-code approaches -- like the ones we used -- assume a close, 1-to-1 \textbf{paraphrase-like mapping} between language and procedure, which misinterpret crucial \emph{framing} and \emph{validation} statements that occurs in abundance in LARC.

%% file: sec_related.tex
\section{Related Works}

\paragraph {Task oriented dialogue systems}
LARC as a dataset belongs to the family of task oriented dialogue systems \cite{nakajima1993study, kelley1984iterative, budzianowski2018multiwoz, wen2016network, andreas2020task}. One can view natural programs in LARC as a single-turn, task-oriented dialogue, where the describer gives a naturalistic instruction with a specific, check-able task in mind. Further, LARC uses the Wizard-of-Oz style of data collection -- leveraging a human interpreter without committing to building a working system -- a common framework to collect data in dialogue systems. LARC differs from these existing datasets mainly in the diversity of its tasks (Section \ref{sec-analysis}) which contain a wide range of abstract concepts rather than being limited to specific domains such as database manipulations \cite{andreas2020task, wen2016network}.

\paragraph{Embodied instruction following}
Embodied instruction following consisting of an embodied agent (often a avatar in a video game) being able to carry out a sequence of commands when prompted with natural language instructions \cite{volum2022craft, suhr2019executing, anderson2018vision, sharma2021skill, wang2016learning}. These commands can often be hierarchical \cite{volum2022craft, sharma2021skill, wang2015building}, which are naturally represented as programs. LARC again differs from these works due to the range of abstract concepts, whereas aforementioned works typically follows a DSL-closed assumption.

As a result of a narrower range of concepts, a paraphrasal strategy that simply translate natural language into code has been fairly successful in prior works that aim to build an instruction following system \cite{andreas2020task, suhr2019executing, wang2016learning}. LARC gives strong evidence that additional grounding strategies need to be modeled to truly capture the richness of natural language instructions (for instance, consider the set of strategies used in Fig \ref{fig:larc_tasks}). 

%% file: sec_conclusion.tex
\section {Conclusion and Future Works}
We present LARC, a \emph{DSL-open} yet \emph{Language-complete} dataset, highlighting the difference of between human-to-human and human-to-machines communications. By annotating successful communications (dataset of \emph{linguistically-tagged-phrases}), we find that humans communicate using a wide range of concepts and communicative strategies, which are difficult to interpret using existing techniques. We hope LARC can help different communities (AI, Programming Language, Cognitive Science, etc) understand and build intelligent, communicative systems. Specifically, we believe that defining concepts upfront (DSL-first) is not scalable. Instead, they should be \emph{learned} and \emph{taught} (by end-users). To fully harness the power of natural language, we need to look beyond the simplistic notion that language having a 1-1 relationship with direct execution, and entertain different communicative strategies \cite{sumers2020learning}. We believe datasets \cite{lachmy2021draw, narayan2019collaborative, suhr2019executing} that share the properties -- namely, DSL-open and language-complete -- are crucial to bridging the gaps between human-human and human-machine communications.
Lastly, it will be beneficial to adapt foundational models \cite{chen2021evaluating, alayrac2022flamingo, ramesh2022hierarchical} -- with some conventional understandings of language, vision, and code -- towards specific domains.

\paragraph{Limitations and Potential Negative Impacts} LARC consists of a single, constrained task format in a highly controlled setting. The long-term goal of this work is to `reverse-engineer' how humans think and communicate, and such systems raise concerns regarding value alignments of users, for instance, non-experts operating safety-critical  equipment using natural language.

%% file: sec_checklist.tex
\section*{Checklist}

\begin{enumerate}

\item For all authors...
\begin{enumerate}
  \item Do the main claims made in the abstract and introduction accurately reflect the paper's contributions and scope?
    \answerYes{}
  \item Did you describe the limitations of your work?
    \answerYes{}
  \item Did you discuss any potential negative societal impacts of your work?
    \answerYes{}
  \item Have you read the ethics review guidelines and ensured that your paper conforms to them?
    \answerYes{}
\end{enumerate}

\item If you are including theoretical results...
\begin{enumerate}
  \item Did you state the full set of assumptions of all theoretical results?
    \answerNA{}
	\item Did you include complete proofs of all theoretical results?
    \answerNA{}
\end{enumerate}

\item If you ran experiments (e.g. for benchmarks)...
\begin{enumerate}
  \item Did you include the code, data, and instructions needed to reproduce the main experimental results (either in the supplemental material or as a URL)?
    \answerYes{See Supplement}
  \item Did you specify all the training details (e.g., data splits, hyperparameters, how they were chosen)?
    \answerYes{See Supplement}
	\item Did you report error bars (e.g., with respect to the random seed after running experiments multiple times)?
    \answerNo{Each synthesis study takes 30 hours and is expensive, and we are not expressly claiming results of the form ``our approach is good'' but only providing suggestions on what may/maynot work}
	\item Did you include the total amount of compute and the type of resources used (e.g., type of GPUs, internal cluster, or cloud provider)?
    \answerYes{See Supplement}
\end{enumerate}

\item If you are using existing assets (e.g., code, data, models) or curating/releasing new assets...
\begin{enumerate}
  \item If your work uses existing assets, did you cite the creators?
    \answerYes{ARC \cite{chollet2019measure}}
  \item Did you mention the license of the assets?
    \answerYes{See Supplement}
  \item Did you include any new assets either in the supplemental material or as a URL?
    \answerYes{See Supplement}
  \item Did you discuss whether and how consent was obtained from people whose data you're using/curating?
    \answerYes{See Supplement}
  \item Did you discuss whether the data you are using/curating contains personally identifiable information or offensive content?
    \answerYes{See Supplement. We do not collect personally identifiable or sensitive information}
\end{enumerate}

\item If you used crowdsourcing or conducted research with human subjects...
\begin{enumerate}
  \item Did you include the full text of instructions given to participants and screenshots, if applicable?
    \answerYes{See Supplement}
  \item Did you describe any potential participant risks, with links to Institutional Review Board (IRB) approvals, if applicable?
    \answerYes{See Supplement}
  \item Did you include the estimated hourly wage paid to participants and the total amount spent on participant compensation?
    \answerYes{See Supplement}
\end{enumerate}

\end{enumerate}


%% file: sec_appendix.tex
\appendix
\newpage
\section{Appendix}

\begin{figure}[h]
\centering
    \includegraphics[width=0.8\textwidth]{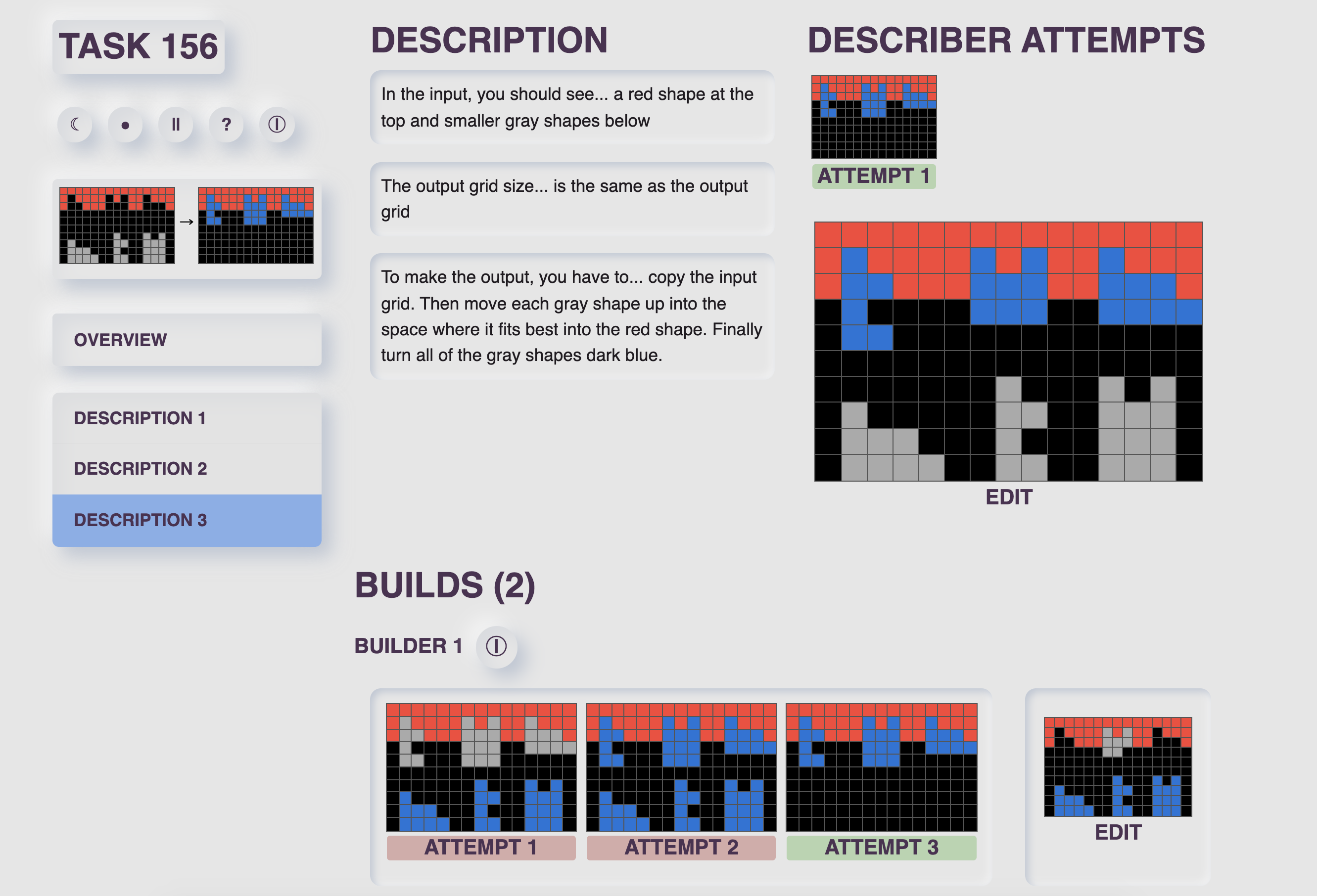}
    \includegraphics[width=0.8\textwidth]{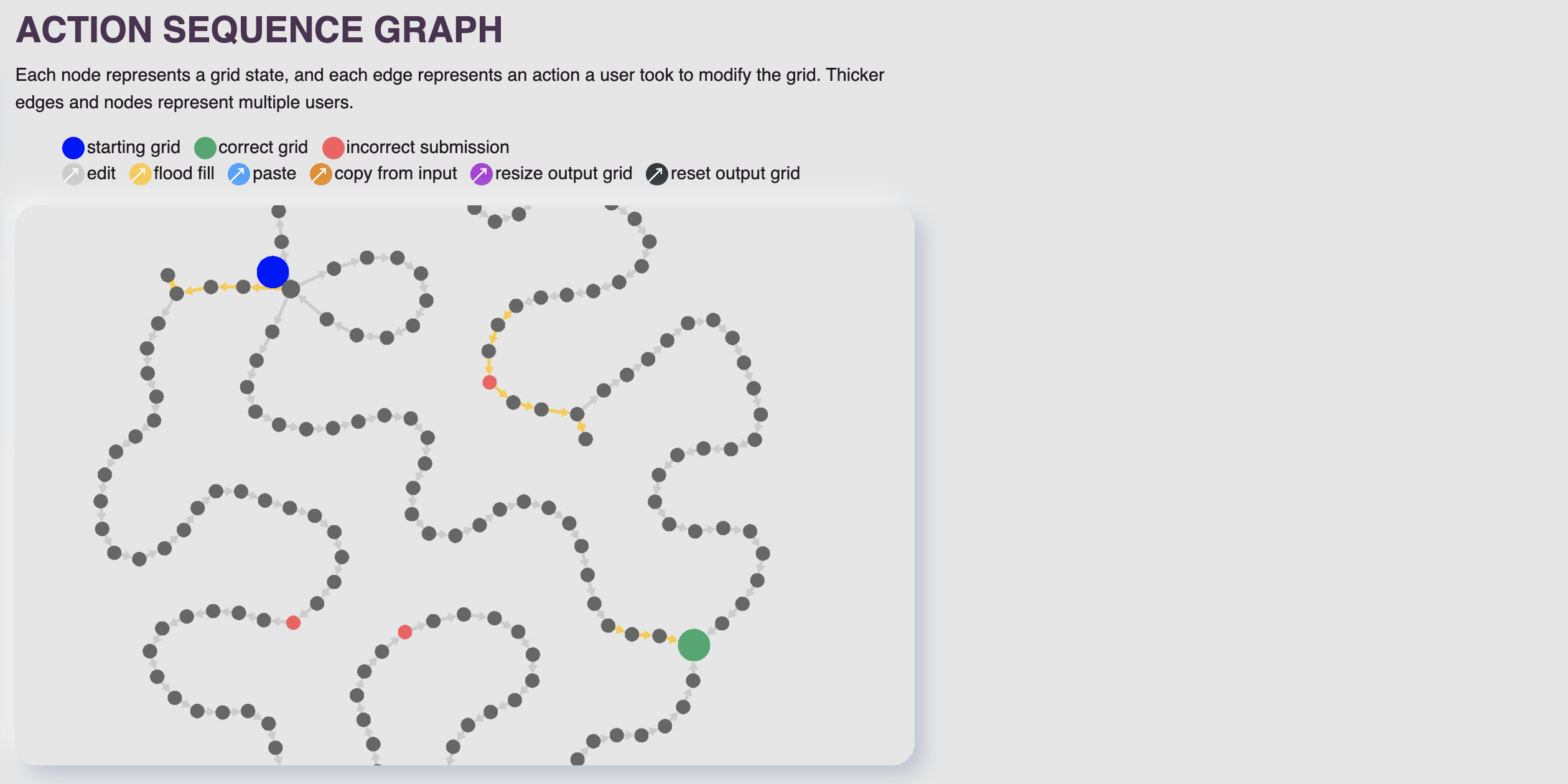}
        
    \caption{The explore interface for task 156 (top). action sequence graph of builder 1 (bot)}
    \label{fig:explorer} 
\end{figure}

The appendix serves as a complimentary document to the paper detailing the data collection process, analysis, and program synthesis. It should be used in conjunction with the following:

\begin{enumerate}
    \item the LARC dataset and its annotation workflow, and bandit algorithm can be found in:
    
    \url{https://github.com/samacqua/LARC}
    
    Which contains the explore gui for the whole dataset entirely in browser (see Fig \ref{fig:explorer}):
    
    \url{https://samacqua.github.io/LARC/explore}

    \item alternatively, one can download the repo and run the explore gui offline:
    \begin{enumerate}
    \item point to the \texttt{LARC} root directory
    \item run \texttt{`python3 -m http.server`}
    \item open \texttt{`localhost:8000/explore/`} in a chrome browser
    \end{enumerate}
    
    \item program synthesis using language codes is at this URL :

    \url{https://github.com/theosech/ec/tree/language-guided_program_synthesis_for_larc}
\end{enumerate}

\subsection{The LARC Explorer GUI}

\newpage
\subsection{Consent Form and Annotation Workflow}
\label{sec:consent_form}

\paragraph{Consent Form}

In this study, you will interpret descriptions of an abstract pattern that you observe in grids. By answering the following questions, you are participating in a study performed by cognitive scientists in [\emph{author institution}]. If you have questions about this research, please contact [\emph{author}] at [\emph{author email}]. Your participation in this research is voluntary. You may decline to answer any or all of the following questions. You may decline further participation, at any time, without adverse consequences. Your anonymity is assured; the researchers who have requested your participation will not receive any personal identifying information about you. By clicking 'I AGREE' you indicate your consent to participate in this study.

\begin{figure}[h]
\centering
    \includegraphics[width=0.84\textwidth]{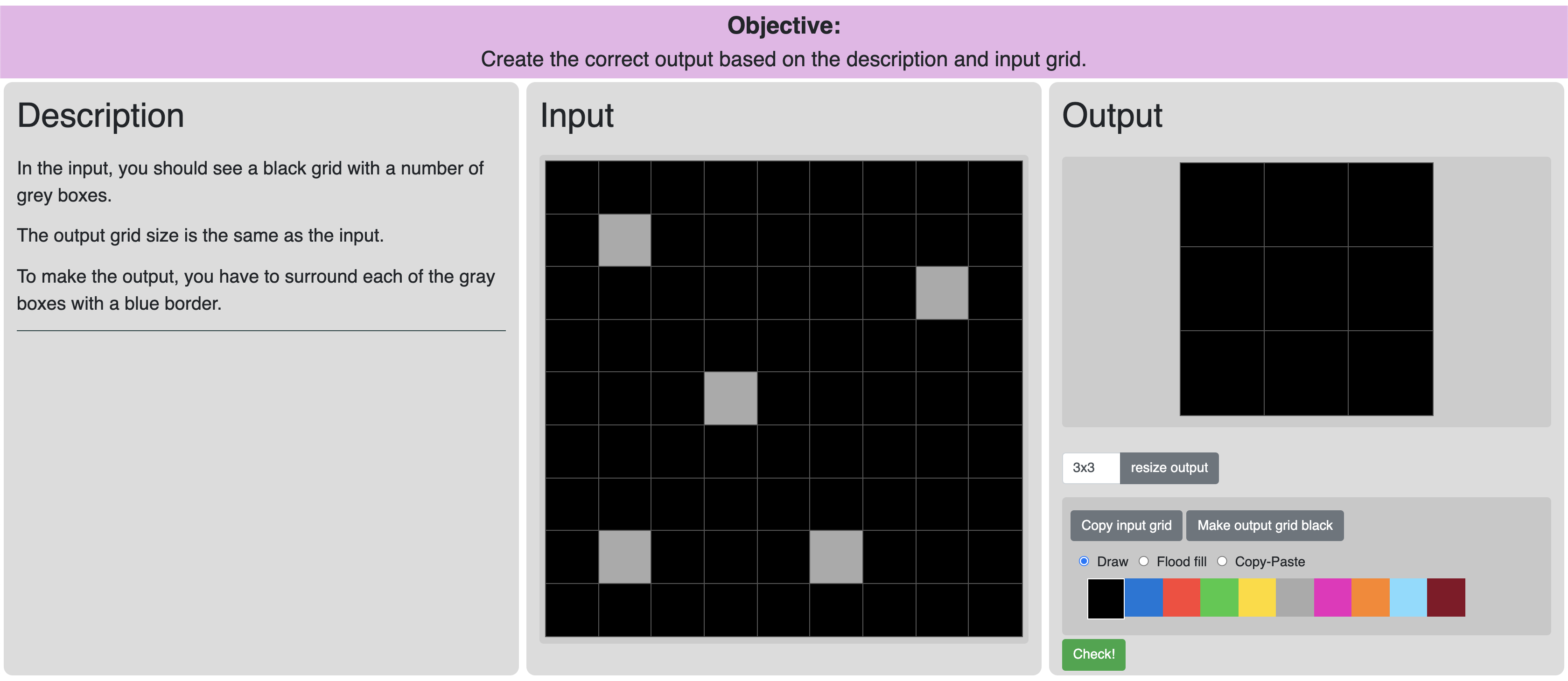}
    \includegraphics[width=0.84\textwidth]{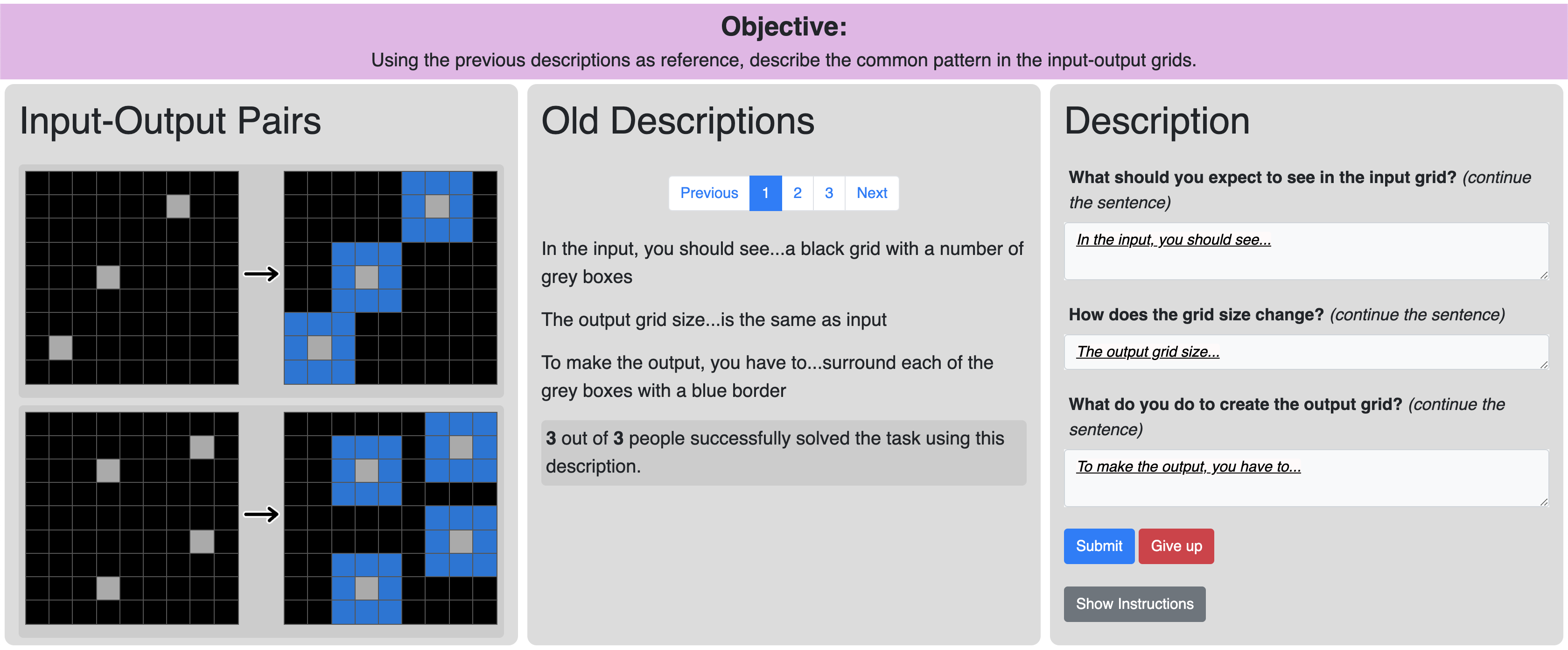}
    \caption{ { \small \emph{A.} The builder interface. \emph{B.} The describer interface. }}
    \label{fig:interface} 
\end{figure}

\paragraph{Annotation Workflow}
Then, the user is given tutorials about communicating ARC tasks, and dynamically assigned a sequence of describe and/or build tasks until they have completed 45 minutes of work. Figure \ref{fig:interface} shows the build and describe interface. For full workflow see \texttt{LARC/collection}.


\newpage
\subsection{LARC Linguistic Analysis Tagging Scheme}
\label{sec:tagging_scheme}
The tagged phrases can be found at \texttt{LARC/dataset/annotated\_phrases.csv}

The phrases were codified by expert coders using a set of 17 binary tags. For each tag, a phrase can either be a positive instance (+) or a negative instance (-) \footnote{marked by 1 and 0 respectively in the csv}. The following table details the tags and coding scheme used:

\begin{center}
\begin{tabular}{ | m{5em} | m{5cm}| m{6cm} | } 
\hline
\textbf{Tag} & \textbf{Description} & \textbf{Examples} \\ 
\hline
\hline
    \textbf{Procedure} & 
    Directly commands the builder to do something; If you were to delete it, the program will fail to execute. & 
    (+) Fill each enclosed hole with yellow
    
    (-) look at the color that form the design in the input. \\ 
\hline
    \textbf{Metaphor} & 
    A metaphor can be an analogy or reference to human common sense knowledge -- e.g. spiral. & 
    (+) A random green pattern
    
    (+) A pattern like a long A \\ 
\hline
    \textbf{Clarification} & 
    A phrase made following a previous statement that attempts to clarify misinterpretations. & 
    (+) Then, copy and paste each colored square in the input grid 4 times -- once in each "quadrant"
    
    (+) (or 5 rows or whatever the number of rows is before it repeats).
    
    (+) Where there's a dark blue square, put orange squares directly above and below it (4 total). \\ 
\hline
    \textbf{Example} & 
    Gives a concrete instance. & 
    (+) The opposite is also true (for example if it is light blue, change to dark red). \\ 
\hline
    \textbf{Array} & 
    Makes a comment about a collection of objects sharing some common property. & 
    (+) Where there's a dark blue square, put orange squares directly above and below it (4 total).
    
(+) Leave the magenta and light blue squares as they are; do not add anything to them if they are present. \\ 
\hline
    \textbf{Validation} & 
    After the builder executes a procedure, check if they got the right answer (i.e. asserts, test-cases, verification, or error handling). & 
    (+) You should end up with all blue boxes touching each other
    
(+) Fill in all of the black boxes to complete the pattern until there are no more black boxes. \\ 
\hline
    \textbf{Loop} & 
    Includes a looping procedure, such as the use of \textit{while}, \textit{for}, \textit{until}, \textit{for each}, or \textit{repeat}. & 
    (+) Continue coloring green until you reach the center of the grid.
    
    (+) Reduce the grid size so that one square is available for each group. \\ 
\hline
    \textbf{Start\_Stop} & 
    Talks about the process or duration of some operations. & 
    (+) start at the upper right corner
    
(+) the red shape needs to move until it is touching the blue cube \\ 
\hline
    \textbf{Conditional} & 
    Of the form \textit{if X then Y}. & 
    (+) If they do not match, make the output square green. \\ 
\hline
    \textbf{Logic} & 
    Includes first-order logic, such as \textit{same}, \textit{and}, \textit{or}, or \textit{not}. & 
    (+) The same size as the input
    (+) You will not use dark blue squares at all
    (-) A 4x4 pattern \\ 
\hline
    \textbf{Framing} & 
    Sets up the problem by offering a particular point of view, defining some objects to be referred to later. & 
    (+) four colored area.
    
(+) 1 or 2 squares filled in with the same color on a black background. \\ 
\hline
\end{tabular}
\end{center}

\begin{center}
\begin{tabular}{ | m{5em} | m{5cm}| m{6cm} | } 
\hline
\textbf{Tag} & \textbf{Description} & \textbf{Examples} \\ 
\hline
\hline
    \textbf{Spacial Relation} & 
    Any reference to a relative position in space to some other component. Positive examples include: under, reaches, touches, angle, outer, downward, parallel, near, after, in between, central, etc. & 
    (+) The red shape next to the blue shape
    
(+) Put yellow inside the green   \\ 
\hline
    \textbf{Physical Interaction} & 
    Any reference to an imaginary force. & 
    (+) The red object falls
    
(+) Blue slides to the left towards red   \\ 
\hline
    \textbf{Contact Transform} & 
    Influence via contact, i.e. any specialized version of physical interaction that involves \textit{at least two objects} and \textit{some type of contact causality}.  & 
    (+) Move X until contact with Y
    
(+) Set X touching Y and turn it the color of Y

(-) Red moves left one square   \\ 
\hline
    \textbf{Affine Transform} & 
    Any reference to a affine transformation over an object, such as rotation, translation, etc.  & 
    (+) Rotate 90 degrees
    
(+) Extend the square into a line  \\ 
\hline
    \textbf{Visual- Graphical Transform} & 
    Any other visual or graphical modification other than a geometric one, such as coloring, flood-fill, or drawing a new shape.  & 
    (+) Make it gray
    
(+) Draw a line  \\ 
\hline
    \textbf{Object Detection} & 
    The localization of a cohesive, bounded object.  & 
    (+) The red shape
    
(+) Move \textbf{it} to the left

(+) The pattern  \\ 
\hline
\end{tabular}
\end{center}

These tags can also be grouped hierarchically into the following categories:

\paragraph{Programmatic:} procedure, array, validation, loop, start\_stop, conditional, logic

\paragraph{Human/Mechanisms for Domain General Communication:} metaphor, clarification, example, framing

\paragraph{Objects and Object Manipulation:} spacial\_relation, physical\_interaction, contact\_transform, geometric\_transform, visual\_graphical\_transform, object\_detection

\input{sec_appendix_dsl}
\input{sec_appendix_synthesis}

\input{sec_appendix_copilot}
\input{sec_appendix_clip}
\input{sec_appendix_synthesis_gpt4}
\input{sec_appendix_bandit}

%% file: sec_appendix_dsl.tex
\newpage
\subsection{THE ATTEMPTED LARC DSL}
\label{sec:larc_dsl}
As LARC is DSL-open, we must first \emph{construct} a suitable DSL before applying (symbolic) program synthesis approaches. Here is our attempt at constructing such a DSL. For each DSL primitives, we also list its corresponding pseudo-annotation comments.
We hand-designed DSL a for the LARC domain consisting of 103 primitives (implemented as a set of polymorphically typed $\lambda$-calculus expressions) intended to be broadly and basically applicable to all tasks on the domain -- the DSL operates over grids of pixels, and contains simple functions designed to repeatedly perform image transformations over pixel grids to produce an output grid. The complete DSL is available at the released code repository; below we provide representative example functions and the accompanying natural language glosses of their behavior used in the \textit{pseudoannotations} generative procedure; as well as sampled program expressions and their generated pseudoannotations.

\begin{longtable}{@{}p{0.5\linewidth}p{0.5\linewidth}@{}}
\toprule
\multicolumn{2}{l}{\textbf{Example DSL Functions and Natural Language Gloss Function Annotations}} \\ \midrule
\textit{DSL Function}                 & \textit{Natural Language Gloss}                            \\ \midrule

blocks\_to\_original\_grid & 'place blocks onto input grid' \\ 
blocks\_to\_min\_grid & 'get the smallest grid containing the blocks' \\ 
first\_of\_sorted\_object\_list & 'get the block with the smallest or greatest value of' \\ 
singleton\_block & '' \\ 
merge\_blocks & '' \\ 
filter\_blocks & 'remove the blocks that have' \\ 
map\_blocks & 'for every block' \\ 
filter\_template\_block & 'find the main block' \\ 
reflect & 'reflect' \\ 
move & 'move' \\ 
center\_block\_on\_tile & 'move block to tile' \\ 
duplicate & 'duplicate' \\ 
grow & 'enlarge' \\ 
fill\_color & 'color the block' \\ 
fill\_snakewise & 'color the block in a snake pattern with' \\ 
replace\_color & 'replace colors' \\ 
remove\_black\_b & 'remove the black background' \\ 
remove\_color & 'remove color from block' \\ 
box\_block & 'get smallest rectangle containing block' \\ 
wrap\_block & 'surround block with' \\ 
filter\_block\_tiles & 'only keep tiles that' \\ 
map\_block\_tiles & 'for each tile of block' \\ 
to\_min\_grid & '' \\ 
to\_original\_grid\_overlay & 'place block on input grid' \\ 
get\_height & 'get height of block' \\ 
get\_width & 'get width of block' \\ 
get\_original\_grid\_height & 'get the height of the input grid' \\ 
get\_original\_grid\_width & 'get the width of the input grid' \\ 
get\_num\_tiles & 'count the number of tiles of the block' \\ 
nth\_primary\_color & 'find the nth most common color' \\ 
is\_symmetrical & 'is the block symmetrical' \\ 
is\_rectangle & 'is the block a rectangle' \\ 
has\_min\_tiles & 'does the block have at least n tiles' \\ 
touches\_any\_boundary & 'does the block touch any edge of the grid' \\ 
touches\_boundary & 'does the block touch the edge' \\ 
has\_color & 'does the block have color' \\ 
is\_tile & 'is the block a tile' \\ 
block\_to\_tile & '' \\ 
get\_block\_center & 'get the central tile of the block' \\ 
map\_for\_directions & 'in every direction' \\ 
find\_same\_color\_blocks & 'find blocks based on shared color' \\ 
find\_blocks\_by\_black\_b & 'find blocks based on if they are separated by the black background' \\ 
find\_blocks\_by\_color & 'find blocks based on if they are separated by the given color background' \\ 
find\_blocks\_by\_inferred\_b & 'find blocks based on if they are separated by the background' \\ 
grid\_to\_block & '' \\ 
split\_grid & 'split the grid in half' \\ 
find\_tiles\_by\_black\_b & 'find the tiles based on if they are separated by the black background' \\ 
is\_interior & 'is the tile in the interior of a block' \\ 
is\_exterior & 'is the tile in the exterior of a block' \\ 
tile\_touches\_block & 'does the tile touch the block' \\ 
tile\_overlaps\_block & 'does the tile overlap the block' \\ 
tile\_to\_block & '' \\ 
extend\_towards\_until & 'extend tile towards a direction until the condition is met' \\ 
extend\_towards\_until\_edge & 'extend tile towards a direction until it touches the edge' \\ 
extend\_until\_touches\_block & 'extend tile towards a direction until it touches the edge' \\ 
move\_towards\_until & 'move tile towards direction until condition is met' \\ 
move\_towards\_until\_edge & 'move tile towards direction until it touches edge' \\ 
move\_until\_touches\_block & 'move tile towards direction until it touches block' \\ 
move\_until\_overlaps\_block & 'move tile towards direction until it overlaps block' \\ 
get\_tile\_color & 'get the color of the tile' \\ 
tiles\_to\_blocks & '' \\ 
filter\_tiles & 'only keep tiles that' \\ 
map\_tiles & 'for every tile' \\ 
overlap\_split\_blocks & 'overlap the split blocks based on colors' \\ 
splitblocks\_to\_blocks & '' \\ 
color\_logical & 'take logical operation on colors using them as true and false' \\ 
land & 'logical operator and' \\ 
lor & 'logical operator or' \\ 
lxor & 'logical operator xor' \\ 
negate\_boolean & 'not' \\ 
map\_tbs & 'for every block in template block scene' \\ 
make\_colorpair & 'make pair of colors' \\ 
north & 'top' \\ 
south & 'bottom' \\ 
west & 'left' \\ 
east & 'right' \\ 
north\_east & 'top right' \\ 
north\_west & 'top left' \\ 
south\_east & 'bottom right' \\ 
south\_west & 'bottom left' \\ 
0 & '0' \\ 
1 & '1' \\ 
2 & '2' \\ 
3 & '3' \\ 
4 & '4' \\ 
5 & '5' \\ 
6 & '6' \\ 
7 & '7' \\ 
8 & '8' \\ 
9 & '9' \\ 
true & '' \\ 
false & '' \\ 
invisible & 'invisible' \\ 
black & 'black' \\ 
blue & 'blue' \\ 
red & 'red' \\ 
green & 'green' \\ 
yellow & 'yellow' \\ 
grey & 'grey' \\ 
pink & 'pink' \\ 
orange & 'orange' \\ 
teal & 'teal' \\ 
maroon & 'maroon' \\ 

\bottomrule
\end{longtable}%

\begin{table*}[h]
\centering
\resizebox{\textwidth}{!}{%
\begin{tabular}{@{}p{0.5\linewidth}p{0.5\linewidth}@{}}
\toprule
\multicolumn{2}{l}{\textbf{Example Sampled Programs and Pseudoannotations}}                        \\ \midrule
\textit{Sampled Program}              & \textit{Natural Language Pseudoannotation}                 \\ \midrule
(lambda (to\_original\_grid\_overlay (remove\_color (grid\_to\_block \$0) yellow) false)) &
  `place block on input grid remove color from block yellow' \\
(lambda (extend\_towards\_until\_edge (block\_to\_tile (grid\_to\_block \$0)) south\_east) true)) &
  `extend tile towards a direction until it touches the edge bottom right' \\
(lambda (blocks\_to\_min\_grid (tiles\_to\_blocks (find\_tiles\_by\_black\_b \$0)) true true)) &
  `get the smallest grid containing the blocks find the tiles based on if they are separated by the black background' \\
                                      &                                                            \\ \bottomrule
\end{tabular}%
}
\end{table*}

Compared to SCONE \cite{long2016simpler}, LARC poses a significantly greater challenge for distant supervision.

\begin{table}[h]
\begin{tabular}{|l|l|l|l|l|}
\hline
              & domain     & dsl size & language kind            & number of instances \\ \hline
LARC          & DSL-open   & 103      & freeform text            & 354                 \\ \hline
SCONE: ALCHEMY & DSL-closed & 24 & step-by-step instruction & 4560                \\ \hline
SCONE: TANGRAMS     &  DSL-closed & 14 & step-by-step instruction & 4989 \\ \hline
SCONE: SCENE     & DSL-closed & 33 & step-by-step instruction&   4402                  \\\hline
\end{tabular}
\caption{Comparison of LARC to SCONE}
\end{table}

%% file: sec_appendix_synthesis.tex
\newpage
\subsection{Supplement to Sec. \ref{sec-synthesis}: Executing Natural Programs}
\label{sec:other_experiments}

\textbf{Enumeration details} For a task, we enumerate from the bi-gram distribution (proposed by the neural model) on a high-powered computing cluster for 720s; and with 24 CPUs in parallel.

\paragraph{Other Models: Neural Sequence Decoder}

We experiment with using a neural sequential decoder which can theoretically capture longer range dependencies. Specifically, we use GRU to decode a program one token at a time. In addition we mask the generated tokens to ensure the generated partial programs are syntactically correct (using the type system). We train using the distant supervision approach exactly as \cite{guu2017language}, with an epsilon-randomized beam search to balance exploiting the current policy and exploring low probability programs under the policy and take gradient steps on discovered programs using the meritocratic parameter update rule. We train using distant supervision on 24 CPUs for 10 hours of wall-clock time on the train split of 200 tasks.
\begin{table}[h]
\begin{tabular}{ccc}
\multicolumn{3}{c}{Neural Sequence Decoder} \\ \hline
   & training tasks discovered & testing tasks solved \\
\hline
IO & 6 / 200 & 2 / 183 \\
IO + NL & 7 / 200 & 0 / 183 \\
NL & - & 0 / 183 \\
\end{tabular}
\label{tab:seq_decoder}
\end{table}
 As we can see, the sequence decoder cannot even recover the 10 seed programs during training, and performs poorly on the testing tasks compared to the bigram model. Consequently, we did not attempt pseudo-annotation on the sequence model.

\paragraph{Other Models: CNN encoding of IO}
We take our best model (IO+NL+pseudo) and additionally condition the neural model with a CNN encoder, rather than leaving it un-conditioned. We find that this model can discover 2 more programs during training and achieves identical outcome to the simpler model without CNN.

\begin{table}[h]
\begin{tabular}{|l|l|l|}
\hline
                 & train  & test   \\ \hline
IO+NL+pseudo     & 21/200 & 22/183 \\ \hline
IO+NL+pseudo+CNN & 23/200 & 22/183  \\ \hline
\end{tabular}
\end{table}


In general, we find that the standard solution to distant supervision, although effective in SCONE, only discovers a few programs in LARC. This finding is unsurprising for the following reasons:
\begin{enumerate}
    \item LARC is DSL-open whereas SCONE is not, thus, there is \emph{no} guarantee that we will discover all LARC programs even if we enumerate an \emph{infinite} number of programs.
    \item In SCONE, every computer program is a sequence of 5 actions that transform the state of the world. A natural language utterance is collected for each of these actions. The language annotation (natural program) is the sequences of these 5 utterances. As a result there a tight alignment from utterance to actions (tokens in the DSL).
    \item SCONE domains have an order of magnitude more tasks to learn from (through distant supervision).
\end{enumerate}
We conclude that collecting simpler, more fine-grained tasks as in SCONE would confer significant benefits to solving LARC, notwithstanding the DSL-open challenge.

%% file: sec_appendix_copilot.tex
\newpage
\subsection{Synthesis with codex}
\label{sec:copilot}

We conduct a exploratory study where we took the 7 tasks solved by the NL+IO specification (in addition to just IO), and see whether github’s co-pilot auto-complete tool (built on codex) can correctly infer the right program using only language as prompt. The prompt is constructed by giving a few hundred of pseudo-annotation - program pairs as context (see A.4), followed by a real NL prompt, and asking co-pilot to rephrase it in the LARC DSL:

\begin{lstlisting}[basicstyle=\small]
# English: not is the block symmetrical color the block maroon
# Program: (lambda (to_min_grid (grid_to_block $0) (negate_boolean  ...

# English: overlap the split blocks based on colors split the grid ...
# Program: (lambda (overlap_split_blocks (split_grid $0 false) (lambda ...

... 400 of such pairs ...

# English: copy only the biggest shape into the output grid
# Program: 
\end{lstlisting}

The top-10 generated candidates are then executed to see if they can generate the correct output for the given task. See Figure below.

\begin{figure}[h]
\centering
    \includegraphics[width=1.0\textwidth]{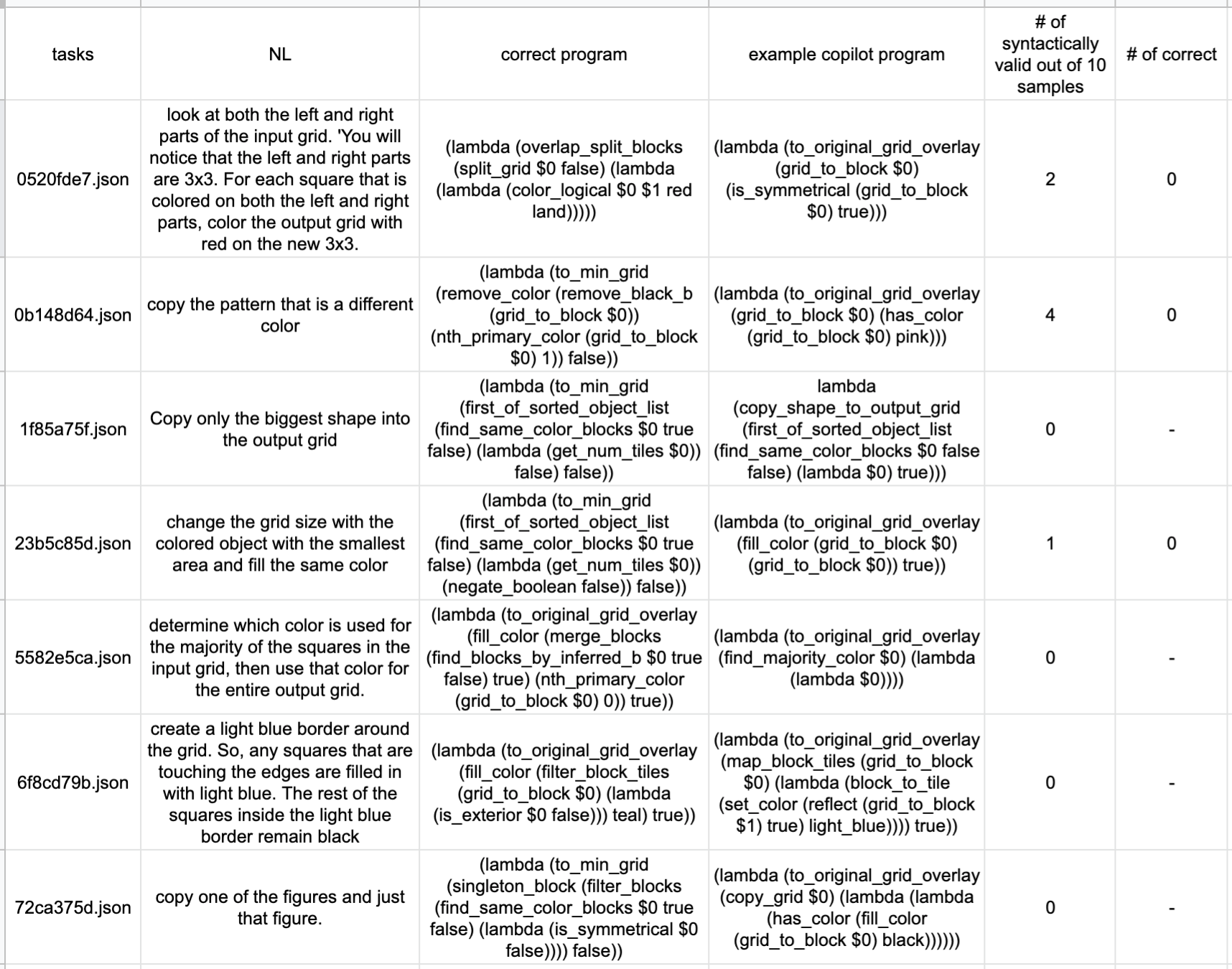}
    \caption{synthesizing programs using copilot (yes, this is a screenshot of a google sheet)} 
\end{figure}

As we can see, while co-pilot suggests programs that look similar to a correct one stylistically, most are syntactically invalid. For instance, it often invents primitives that do not even exist in our DSL, such as “copy\_shape\_to\_output\_grid”. Further, none of the syntactically correct programs can produce the intended output either. This is to be expected, as we use a DSL that has not been seen before in any existing corpus of code (on github), and we should not expect codex to perform well naively. Taking a general model (such as codex) and specializing it to a specific context (LARC) will be exciting future research.

%% file: sec_appendix_clip.tex
\newpage
\subsection{Description Pairing Study with Clip}
\label{sec:clip}

We conduct a exploratory study whether the CLIP model \cite{radford2021learning}, which computes a similarity score between image and captions, can perform the simple task of correctly pair a test input grid with its corresponding description in LARC. Performance on this simple binary classification task is a reasonable upper-bound on how large pre-trained models (such as CLIP, Flamingo \cite{alayrac2022flamingo}, or DALLE \cite{ramesh2022hierarchical}) would work on LARC out of the box. 

\begin{figure}[h]
\centering
    \includegraphics[width=0.8\textwidth]{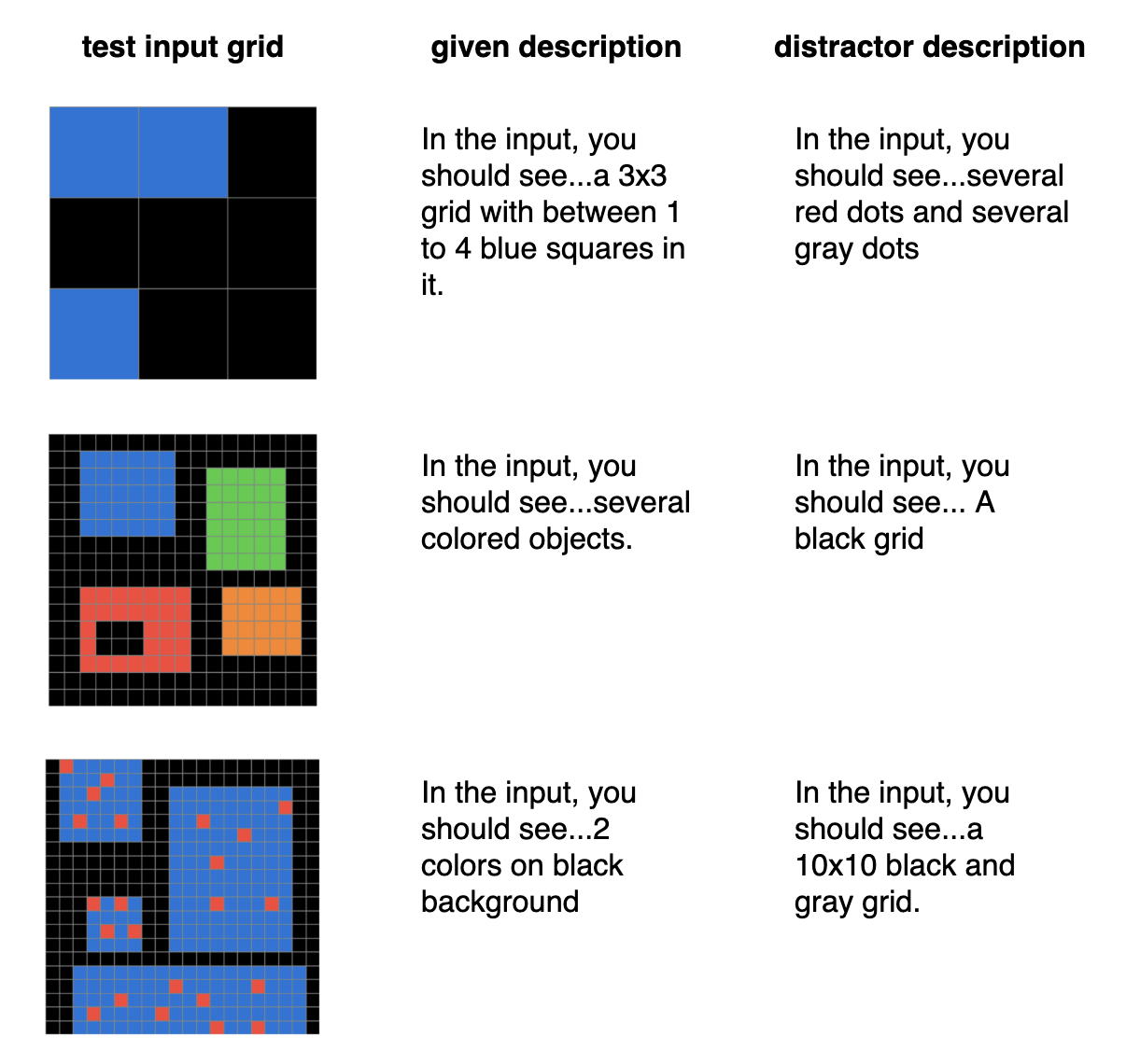}
    \caption{a few instances of the pairing task} 
    \label{fig:pair-test}
\end{figure}

Specifically, we sampled 1000 instances of $(test\_input\_grid, paired\_description, distractor\_description)$ where the paired description comes from the same LARC task, and the distractor description is randomly chosen from a different task. See Figure \ref{fig:pair-test}.

We find that CLIP was able to correctly pair the input-grid with its description (having a higher similarity score than the distractor) 64\% of the times (randomly guessing will have 50\%). To understand how it is making the pairing, we replaced all occurrences of a color word (such as `red' or `black') with the dummy word `COLOR'. After this substitution, the performance drops to 56\%. In conclusion, there are certainly values in using a large pre-trained model that builds a joint representation between language and images. However, achieving only 64\% accuracy on a extremely simplified, binary classification task, where most of the benefits comes from low level concepts such as color, motivates further research endeavours towards actually solving LARC using pure neural approaches -- i.e. \emph{generating} a correct, pixel perfect output grid from input-grid and language alone.

%% file: sec_appendix_synthesis_gpt4.tex
\newpage

\section{Executing Natural Programs Using GPT4}
In this preliminary study, We encode all 354 successfully completed LARC tasks as prompts and attempted to use GPT4 (no vision, language only) to solve them by generating the output directly in the manner of neural program induction \cite{graves2014neural}. Of these tasks, 109 can be solved via prompting.

\subsection{Difference between Program Synthesis and Program Induction}

\paragraph{Neuro-Symbolic Program Synthesis}
In (neuro-symbolic) Program synthesis \ref{sec-synthesis}, the ultimate goal is to produce a valid \emph{computer program} to be executed on a symbolic interpreter (e.g. python), which can be run on additional inputs to produce valid outputs. 

\textbf{pros} (1) cheap to perform inference (i.e. execute python code) if a program is found. (2) strong generalization -- correctness on test-case likely to guarantee in correctness on unseen inputs.

\textbf{cons}  difficult to set up, as evident in Section \ref{sec-synthesis}.

\paragraph{Neural Program Induction}
On the other hand, in neural program induction, the
neural net itself acts as an interpreter, which carries out the computation on additional inputs. The most prevalent form of neural program induction takes the form of prompting \cite{wei2021finetuned}, where the task is encoded as a prompt -- a sequence of tokens, and and LLM decodes the answer using this sequence as context.

\textbf{pros:} trivial to set up -- just put in the prompt and let LLM generate the output as a sequence.

\textbf{cons:} (1) no guarantee if the execution consistently generalizes to unseen inputs. (2) expensive -- each additional new input requires a call to an LLM, which is more expensive than a call to python.

\begin{figure}[h]
\centering
    \includegraphics[width=0.95\textwidth]{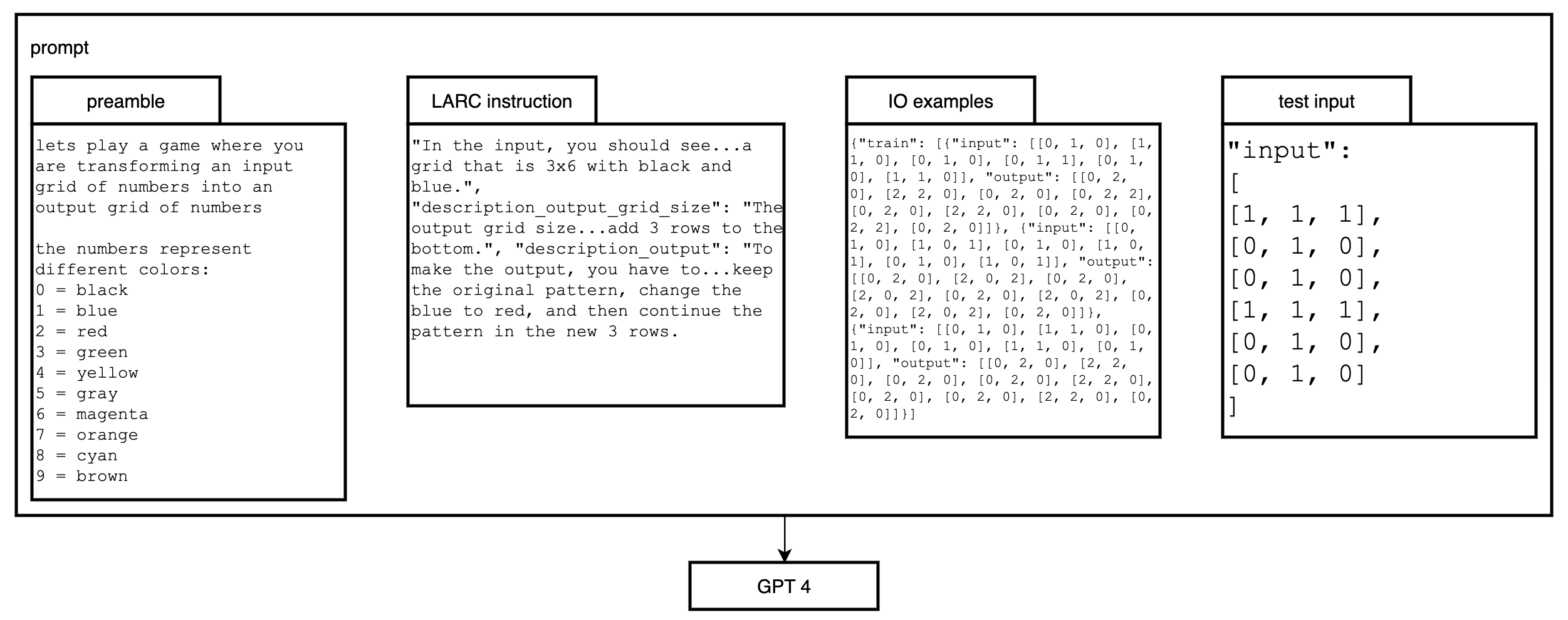}
    \caption{prompt for running LARC tasks on gpt4. For the IO+NL setting, both the instruction and the IO examples were given in the prompt. For IO-only, the LARC instruction is omitted, and for NL-only, the IO examples were omitted} 
    \label{fig:gpt4}
\end{figure}

\paragraph{The Prompts} 
An outline of the prompts considered is shown in Figure \ref{fig:gpt4}. We consider 3 kinds of prompts same as the conditions in Section \ref{sec-synthesis}: IO-only, NL-only, IO+NL. Each prompt is queried to the gpt4 API (temperature = 1.0), and costed roughly 350 dollars total.

\begin{figure}[h]
\centering
    \includegraphics[width=0.8\textwidth]{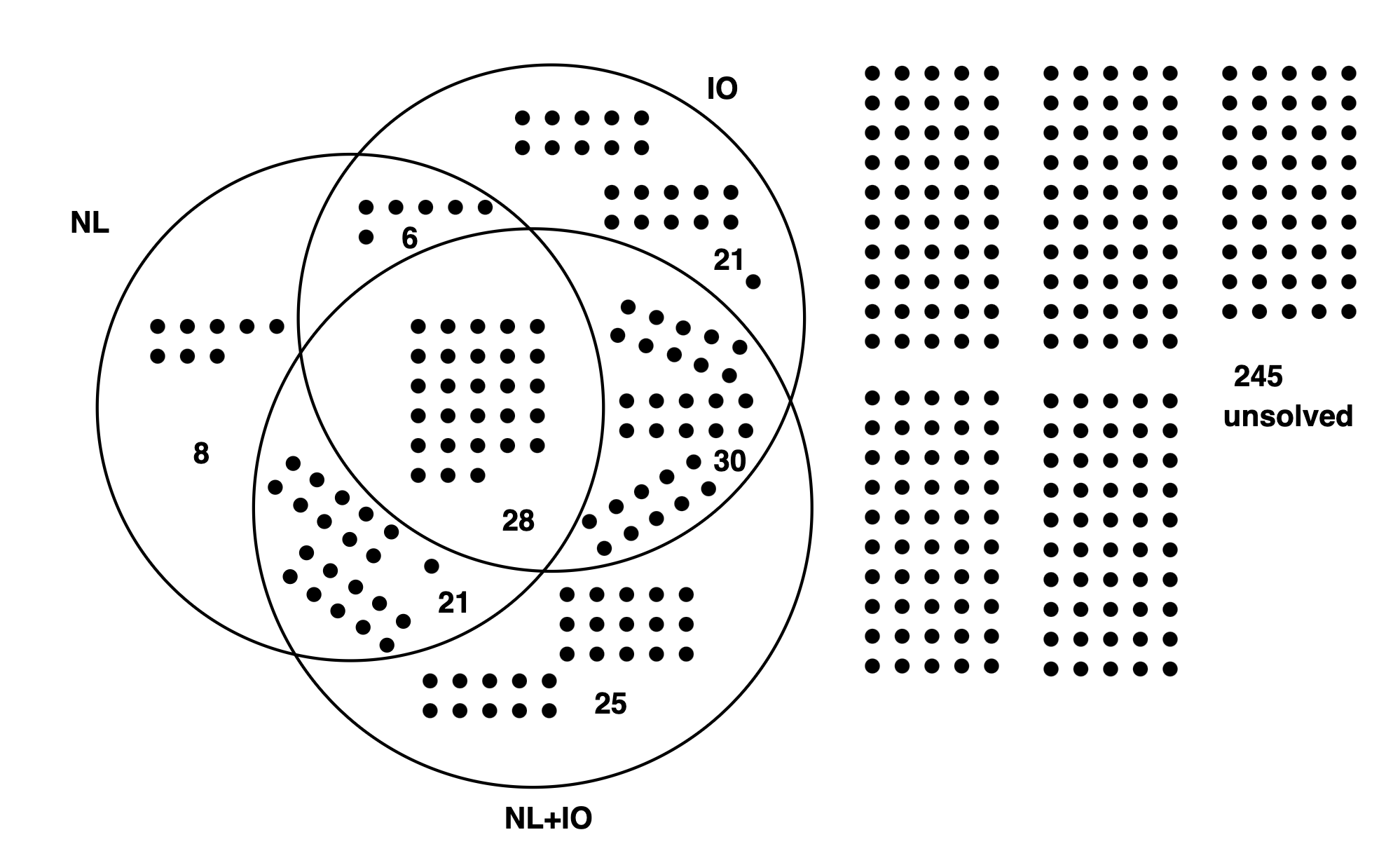}
    \caption{a venn diagram of all the tasks ran on gpt4} 
    \label{fig:gpt4-venn}
\end{figure}

\paragraph{Result}
The result is summarized in a venn diagram (Figure \ref{fig:gpt4-venn}). We find that: \textbf{there are many unsolved tasks}, of 354 successfully completed LARC tasks (by human), 245 remains unsolved by gpt4. \textbf{NL+IO performs best} corroborating out findings in Section \ref{sec-synthesis}, the best performing natural-program leverages both natural language \emph{and} input-output examples. 

GPT4 has solved 109 tasks, outperforming the program synthesis results in Section \ref{sec-synthesis}. However, unlike neuro-symbolic approaches, it is unknown if the prompt would continue to generalize on additional unseen inputs.

\begin{figure}[h]
\centering
    \includegraphics[width=0.8\textwidth]{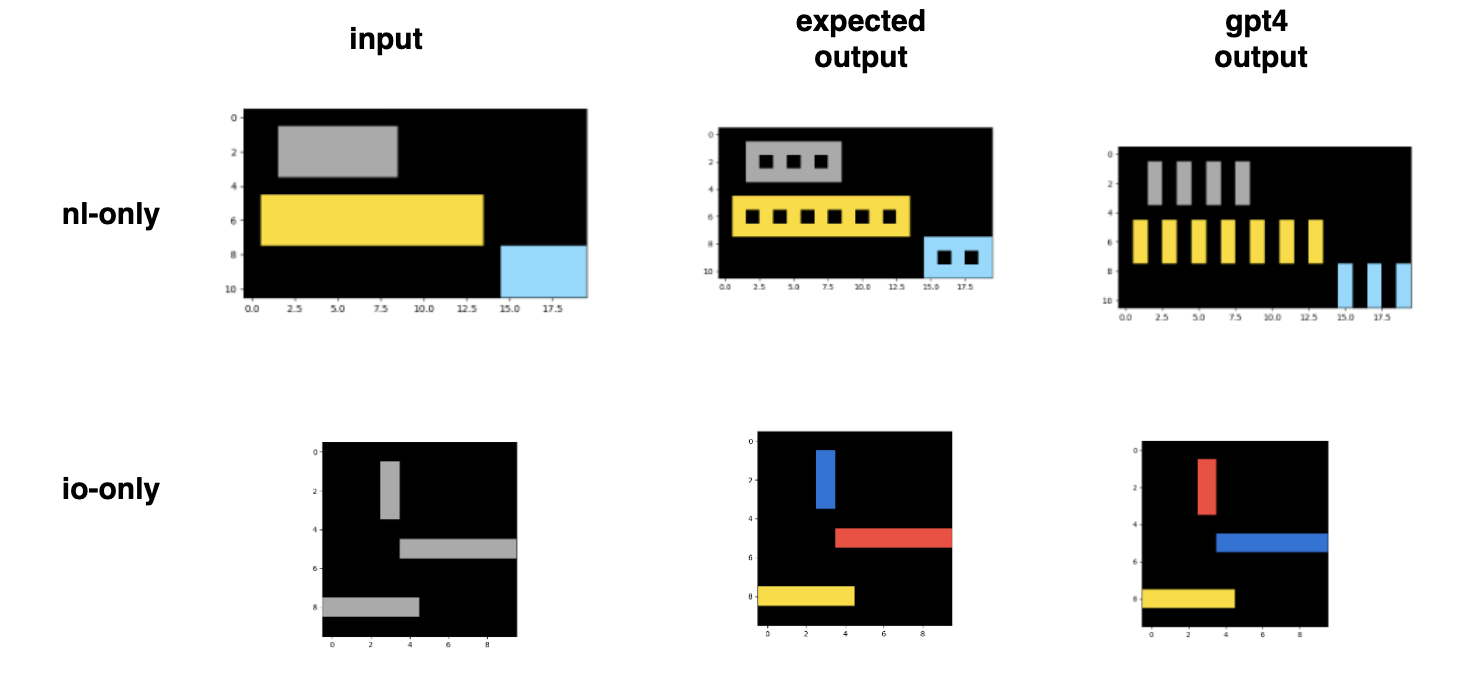}
    \caption{Two tasks that were correct in the IO+NL condition, but omitting either IO or NL cause the task to fail} 
    \label{fig:gpt4-qualitative}
\end{figure}

We found that IO and NL contributes to different aspects of an LARC task. See Figure \ref{fig:gpt4-qualitative}. The top task become incorrect in the NL-only condition, as with natural language alone, the notion of ``drilling a hole'' is ambiguous. The bottom task become incorrect in the IO-only condition, as all GPT4 can infer from the IO examples were that a combination of blue, yellow, and red were used. However, given the natural language of LARC that explicitly instructs the listener to color the shapes in decreasing size red, yellow, and blue, it is possible for GPT4 to apply this given rule.

To re-iterate, the ARC task consists of 2 components, a program (a rule) must be first \emph{synthesized} from the given IO examples, this program must then be \emph{executed} on an additional input. LARC solved the first problem, by using annotators to first solve for the program, thus, GPT4 only had to apply the rule, without having to infer it in the first place.

All the GPT4 results can be found here (\url{https://github.com/evanthebouncy/larc_gpt4}). 

We would like to thank Simon Alford for his pdf generation code that summarizes the results.

%% file: sec_appendix_bandit.tex
\newpage

\section{Appendix : multi-bandit, infinite-arm, best-arm identification}
\label{sec:bandit_algo}

Imagine there are N different mAgIcAl casinos, where each has an infinite number of slot machines (arms). While each individual arm has its own probability $p$ (Bernoulli) of generating an outcome of either 0 or 1, the arms are related to each other depending on the casinos they belong to. Some casinos are easier than others, in a sense that for some, it is easier to find a ``good'' arm whereas for others, most arms will have a small chance of success. Moreover, each casino $i$ has one (or multiple) best arm, whose probability of generating a 1 is $p_i^*$. Your job is to identify the best arm within each casino. This is in essence the multi-bandit, infinite-arm, best-arm identification problem. 

You can take observations in the casinos, where each observation involves selecting a casino, and trying one of its arms (either one of the arms you already tried, or trying a new one out of its infinite possibilities), observing an outcome of either 0 or 1. We seek an online algorithm that, given any observation budget, propose a set of N arms. Let $p_1 \dots p_N$ denote the ground-truth Bernoulli parameters of the proposed arms. We seek to minimize the following regret:

$$
    L = \sum_i (p_i^* - p_i)
$$

Where each term $p_i^* - p_i$ is the ``gap'' between the proposed arm and the best arm in a given casino.

\subsection{Application to LARC}
Our goal is to collect a working natural program for each of the 400 ARC tasks.
Natural programs are difficult to collect, because it involves both: 1) obtaining a natural program from a describer and 2) validating this natural program by having a builder build from it. Thus, rather than exhaustively studying each task to estimate its difficulty, we are content with just getting a ``good enough'' natural program for each task. In another words, given a certain annotation budget, we want to find a single good natural program for each of the 400 tasks. 

If we take the 400 tasks as 400 casinos, then each casino would have an intrinsic difficulty, which corresponds to how easy it is to communicate a particular task. Within each task, there are an infinitely many possible natural programs (i.e. all natural language strings), which correspond to the infinite-arm aspect. For each task, we are interested in finding as good of a description as we can, which correspond to the best-arm identification aspect.

\begin{figure*}[h]
\centering
    \includegraphics[width=0.3\textwidth]{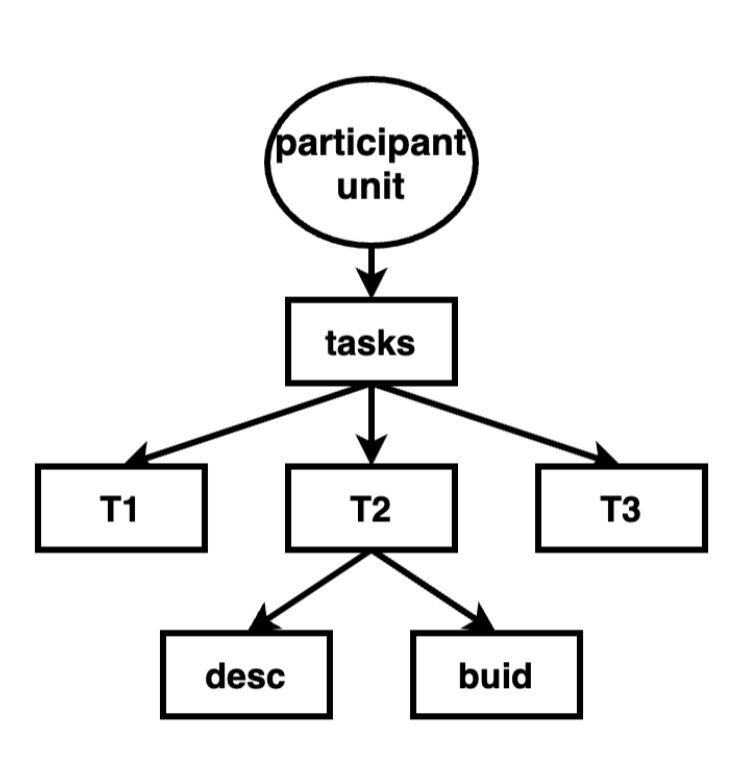}
    \caption{How a ``unit'' of a participant's time can be utilized}
    \label{fig:turk_bandit} 
\end{figure*}

Specifically, we are seeking an online algorithm that \emph{at any budget} can propose a set of natural programs, and this set of proposed programs should improve with added budget (budget here is synonymous with total participants' time). To use the bandit algorithm in conjunction with the annotation process, we divide the 45 minutes of a participant's time into several ``units'' of participation, where each unit can be assigned to one of two jobs: 1) The participant can either give a new description to an ARC task, then immediately build from it (in the form of describer verification) or 2) The participant can be given an existing description of a task, and build from it to to assess if it is a good description. See Figure \ref{fig:turk_bandit}. We estimate how many minutes would this particular unit take, and dynamically allocate additional units until the full 45 minutes are exhausted.

\subsection{Reinforcement Learning Formulation}
\begin{figure*}[h]
\centering
    \includegraphics[width=0.9\textwidth]{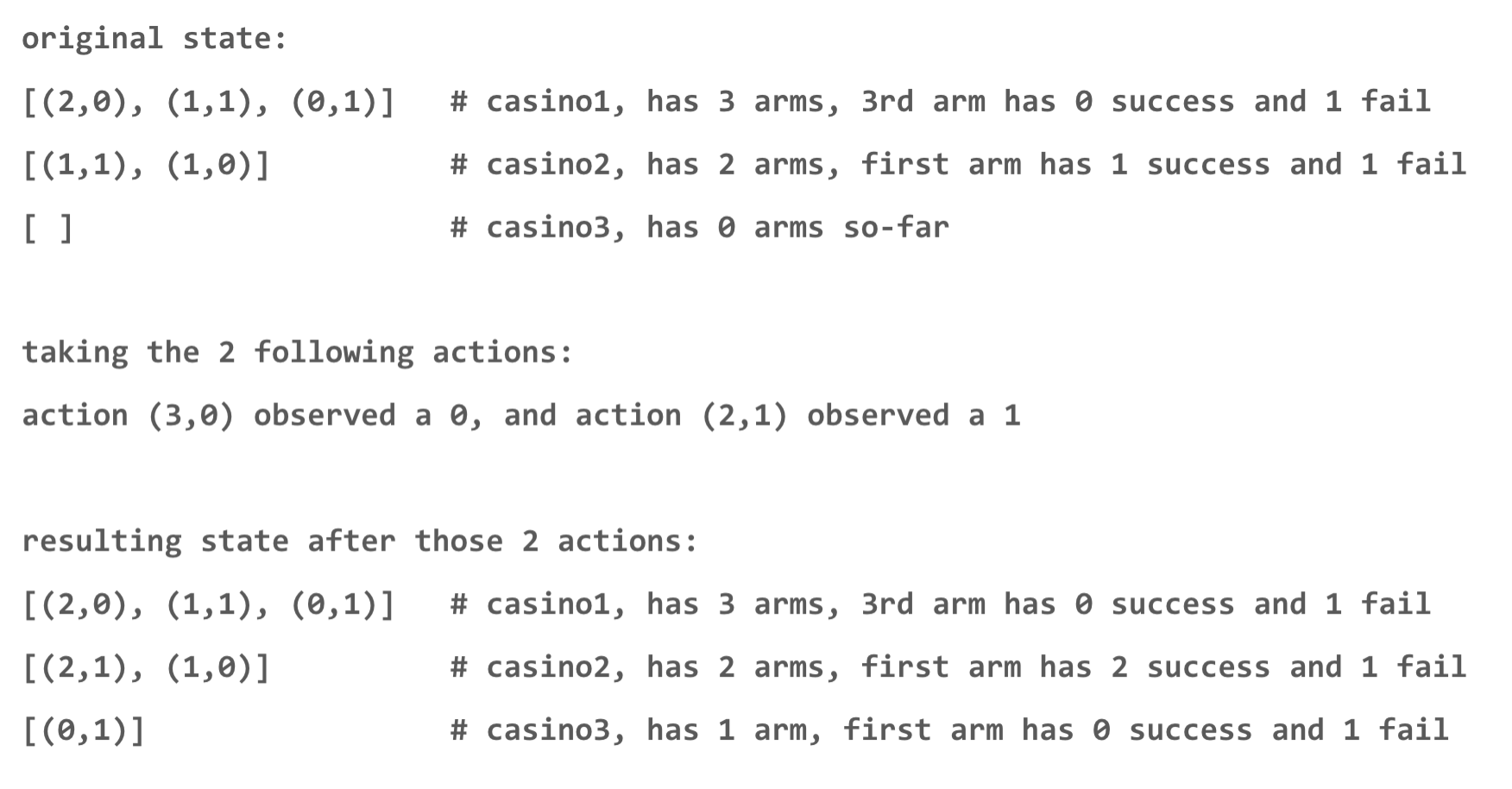}
    \caption{an example transition where there are 3 casinos}
    \label{fig:bandit_transition} 
\end{figure*}

A great way to formalize a bandit problem is casting it as an instance of a Markov Decision Process:

A \textbf{state} consists of all the observations (the 0, 1 outcomes) on all the arms thus far. Let there be $N$ bandits/casinos, then the observation is a collection of all casinos' outcomes $C_1 \dots C_N$ where for each casino $C_i$, we have observation for its $K$ arms that we already sampled: $c_i^1 \dots c_i^K$. Each arm's observation, $c_i^j$ is simply a tuple $(A,B)$ where $A$ denotes the number of 1s observed from arm $c_i^j$ and $B$ denotes the number of 0s. Thus, the space of observation is $O(N \times K \times (A+B))$. See Figure \ref{fig:bandit_transition}.

There are two kinds of \textbf{actions} -- the \emph{arm-selection} action, and the \emph{best-arm-proposal} action. Arm selection consists of a tuple $(i,j)$ where $i$ selects a casino, and $j$ selects from which of the arms within that casino to sample an additional observation. We will use $j = 1 \dots K$ to denote sampling from the $K$ arms within a particular bandit $i$, and use $j = 0$ to denote sampling a \emph{new} arm from bandit $i$. When the interaction budget is exhausted, the agent must make a best-arm-proposal action, in which the agent picks one sampled arm from each casino to be calculated in the regret. For arm proposal, we use a simple heuristic that selects the arm with the highest estimated mean using a beta distribution with (1,1) prior. For the remainder of this section, \textbf{action} will refer exclusively to arm-selection.

\paragraph{Transition} modifies the \textbf{state} to include the new observation. See Figure \ref{fig:bandit_transition}.

\paragraph{Reward} is the sum of the Bernoulli parameters for the set of proposed arms. $p_1 + \dots + p_N$.

\subsection{A Heuristically Defined Agent}
To the best of our knowledge, there is no bandit algorithm that address the specific bandit problem we are solving. However \cite{wang2008infinitely} solves the infinitely many armed bandit problem for a single bandit, where they explicitly model the difficulty of the underlying bandit. We take their algorithm as inspiration. Note that \cite{wang2008infinitely} prescribe a solution to the regret-minimization problem, which is not exactly best-arm-identification. However, in the limit, the two are equivalent as minimizing regret is equivalent to finding the optimal arm. We will first state the result of \cite{wang2008infinitely}, which applies to the case of a single casino/bandit, then extend it to the case of multi-bandit.

\paragraph{arm selection}
Suppose we know that we want to generate an action in casino $i$. \cite{wang2008infinitely} proposed the following rule for selecting which arm to interact with. Let $\beta$ be the difficulty parameter of the task, defined as: $P(p^* - p_j < \epsilon) = \Theta (\epsilon ^ \beta)$. Which is to say, if you were to sample a new arm with ground truth parameter $p_j$, the probability that this arm lies within $\epsilon$ of the optimal arm, is approximately $\epsilon ^ \beta$. For instance, if $\beta = 1$, the task is very difficult as $\epsilon ^ 1$ is a tiny number, meaning it is almost impossible for you to sample an arm $p_j$ that is $\epsilon$ close to optimum. Conversely, if $\beta = 0$, the task is very simple, as $\epsilon ^ 0 = 1$, so any arm you sample will be optimal.

\cite{wang2008infinitely} states that, if you let $M$ be the total number of observations on a bandit, and $K$ be the total number of arms currently sampled, if $K \leq M^\beta$, then you should sample a new arm. Otherwise, you should perform the standard UCB algorithm on the set of existing arms. In our bandit RL environment, $M$ and $K$ are well defined, but how do we estimate $\beta$? We use the following heuristic to estimate difficulty: Let $j$ be the best arm in the current casino w.r.t. its sampled mean $\tilde{p_j}$, then we define $\beta = 1 - \tilde{p_j}$. For instance, if the best arm has a sampled mean of 0.9, then we are in an ``easy'' casino, and the difficulty will be $1-0.9=0.1$, which is fairly close to 0, implying we should \emph{NOT} be sampling new arms, as the best arm we have currently is likely to be good. Conversely, if the best arm has a sampled mean of 0.1, then we are in a ``difficult'' casino, where we stand a better chance of finding a good arm by sampling more arms.

\paragraph{casino selection} To adopt the infinitely-many arm algorithm to a multi-bandit setting, we use the following heuristic: selecting the casino where we have the least information about $p^*$ of a casino. In practice, we rank all $K$ arms based on their sampled mean, and take the top-half of the arms, and aggregate a beta distribution of the total number of 1s and 0s of these arms, and use the variance of the beta distribution as a proxy for uncertainty. For instance, if a casino whose top-half arms have in total many observations, and most of them are 1s, then we are certain about its $p^*$. Conversely, if a casino whose top-half arms have few observations, and it is an even split of 1s and 0s, we are unsure of its $p^*$.

\subsection{Simulated Evaluation}
With both arm selection and casino selection, we have a functioning agent. We can evaluate this agents' performance against several baseline agents in the bandit RL environment to verify that it is indeed more efficient. We consider the following baseline agents, \textbf{rand} is the random agent that select an action at random, \textbf{tile} is the agent that tries to evenly spread out the observation budget, \textbf{tile-inf} is the agent that uses the infinitely many arm algorithm, and tries to spread the budget evenly across casinos, \textbf{cas-inf(ours)} is the agent that selects the casino using uncertainty of $p^*$, and use infinitely many arm algorithm.

\begin{figure*}[h]
\centering
    \includegraphics[width=0.6\textwidth]{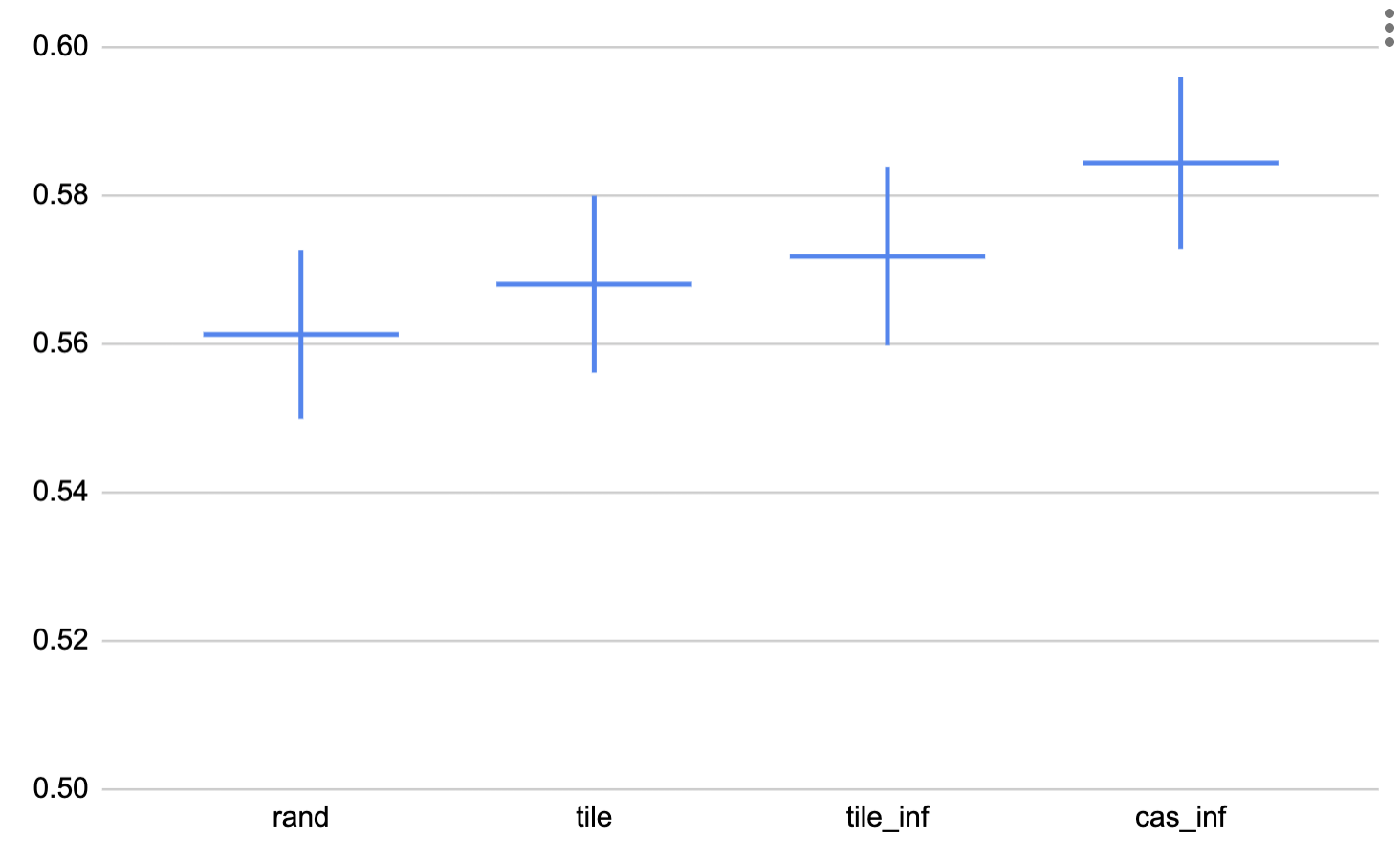}
    \caption{performance of various bandit policies, of 100 casinos and a budget of 600, averaged across 100 repetitions. horizontal bar is average, whiskers indicate standard deviation}
    \label{fig:bandit_result} 
\end{figure*}

The algorithms performance over 100 casinos with a total of 600 interaction budgets is in Figure \ref{fig:bandit_result}

As one can see, for the simulated environment, which makes several simplifications, such as not taking in the generation aspect of description making, and modeling difficulty of a casino as a truncated gaussian, our proposed bandit algorithm out-performs the other baselines. The implementation of the bandit environment and the bandit policies can be found at \texttt{LARC/bandit}